\documentclass{article}

\PassOptionsToPackage{numbers, compress}{natbib}
 \usepackage[main, final]{neurips_2026}
\usepackage[utf8]{inputenc} % allow utf-8 input
\usepackage[T1]{fontenc}    % use 8-bit T1 fonts
\usepackage{hyperref}       % hyperlinks
\usepackage{url}            % simple URL typesetting
\usepackage{booktabs}       % professional-quality tables
\usepackage{amsfonts}       % blackboard math symbols
\usepackage{nicefrac}       % compact symbols for 1/2, etc.
\usepackage{microtype}      % microtypography
\usepackage{xcolor}         % colors
\usepackage[normalem]{ulem}
\useunder{\uline}{\ul}{}
\usepackage{pifont}
\usepackage{multirow}
\usepackage{amsfonts}
\usepackage{float}
\usepackage{graphicx}
\usepackage{algorithm}
\usepackage{algpseudocode} % 新版伪代码环境
\usepackage[table]{xcolor}
\usepackage{enumitem}

\usepackage{amsmath,amssymb}
\usepackage[nameinlink,capitalize]{cleveref}
\usepackage{subcaption}  % 处理子表格
\usepackage{adjustbox}  % 处理表格大小调整
\usepackage{booktabs}
\definecolor{darkgreen}{RGB}{0,150,0}  % 暗绿色 (RGB: 深绿)
\definecolor{darkred}{RGB}{150,0,0}   % 暗红色 (RGB: 深红)
\definecolor{red_2}{RGB}{192,0,0}  % 暗红色 (RGB: 深红)
\definecolor{darkblue}{RGB}{0,0,150} 
\definecolor{darkpurple}{RGB}{110, 0, 150}
\definecolor{darkorange}{RGB}{255, 140, 0}
\definecolor{Baselineblue}{RGB}{0, 90, 190}     % Baseline 网络（深蓝）
\definecolor{hotpink}{RGB}{255,105,180}
\definecolor{hermesorange}{RGB}{255,127,0}
\definecolor{orange}{RGB}{234,112,12}
\definecolor{palegreen}{RGB}{120,148,64}

\definecolor{bestbg}{HTML}{F4B6B6}    % deeper soft red
\definecolor{secondbg}{HTML}{BFD8FF}  % deeper soft blue
\definecolor{fdepbg}{HTML}{F3F4F6}    % very light gray for FDEP rows
\newcommand{\best}[1]{\cellcolor{bestbg}\textbf{#1}}
\newcommand{\second}[1]{\cellcolor{secondbg}#1}

\definecolor{bestbg_deep}{HTML}{F2A6A6}
\definecolor{secondbg_deep}{HTML}{AFCFFF}

\newcommand{\gain}[1]{\!\textcolor[HTML]{C00000}{\tiny\textbf{(#1)}}}

\title{Rethinking Infrared Small Target Detection: A Foundation-Driven Efficient Paradigm}

\author{\textbf{Chuang Yu}$^{1,2,6}$ \,
\textbf{Jinmiao Zhao}$^{1,2}$ \,
\textbf{Yunpeng Liu}$^{1}$\footnotemark[2] \,\,\,  
\textbf{Yaokun Li}$^{3}$ \, 
\textbf{Xiujun Shu}$^{2}$  \\
\textbf{Yuanhao Feng}$^{4}$ \,
\textbf{Bo Wang}$^{5}$  \,
\textbf{Xiangyu Yue}$^{6}$\footnotemark[2] \\
\\
$^{1}$Shenyang Institute of Automation, Chinese Academy of Sciences \quad \\
$^{2}$University of Chinese Academy of Sciences \quad
$^{3}$Sun Yat-sen University \quad
$^{4}$Peking University  \quad \\
$^{5}$Northwestern Polytechnical University   \quad
$^{6}$MMLab, The Chinese University of Hong Kong   \quad
}

\begin{document}

\renewcommand{\thefootnote}{\fnsymbol{footnote}} 
\footnotetext[2]{\ Corresponding author: \texttt{ypliu@sia.cn} and \texttt{xyyue@ie.cuhk.edu.hk}.}

\maketitle

\begin{abstract}
While large-scale visual foundation models (VFMs) exhibit strong generalization across diverse visual domains, their potential for single-frame infrared small target (SIRST) detection remains largely unexplored. To fill this gap, we first systematically introduce frozen VFM representations into the SIRST task and propose a \textbf{\underline{F}}oundation-\underline{\textbf{D}}riven \underline{\textbf{E}}fficient \underline{\textbf{P}}aradigm (\textbf{FDEP}), a general and efficient framework compatible with diverse VFMs and SIRST networks, which improves detection accuracy without additional VFM-related inference overhead. Specifically, a Semantic Alignment and Modulated Fusion (SAMF) module is designed to achieve dynamic alignment and deep fusion of the global semantic priors from VFMs with task-specific features. Meanwhile, to avoid the inference-time overhead introduced by VFMs, we propose a Collaborative Optimization-based Implicit Self-Distillation (CO-ISD) strategy, which enables implicit semantic transfer between the main and lightweight branches through parameter sharing and synchronized backpropagation. In addition, to unify the fragmented evaluation system, we construct a Holistic SIRST Evaluation (HSE) metric that performs multi-threshold integral evaluation at both pixel-level confidence and target-level robustness, providing a stable and comprehensive basis for fair model comparison. Extensive experiments demonstrate that the SIRST detection networks equipped with our FDEP framework achieve state-of-the-art (SOTA) performance on multiple public datasets. Our code is available at \href{https://github.com/YuChuang1205/FDEP-Framework}
{\textcolor{hotpink}{\textbf{https://github.com/YuChuang1205/FDEP-Framework}}}.
\end{abstract}

\section{Introduction}
\label{sec:intro}

Single-frame infrared small target (SIRST) detection is a key technology for infrared search and tracking systems~\citep{yu2025easy,gao2025bio,zhang2024single,yuan2023thermal} and has been widely applied to target tracking~\citep{ying2025infrared,tu2022rgbt,zhao2024refined}, maritime assistance~\citep{zhang2022exploring,zhao2024infrared,han2022kcpnet}, and early warning~\citep{li2018robust,xu2023multiscale,najafi2024high}. However, current SIRST detection remains challenging due to complex backgrounds, scarce target features, and severe class imbalance.

\begin{figure}[t]
  %\captionsetup{skip=2pt}
  \centering
   \includegraphics[height=4.8cm]{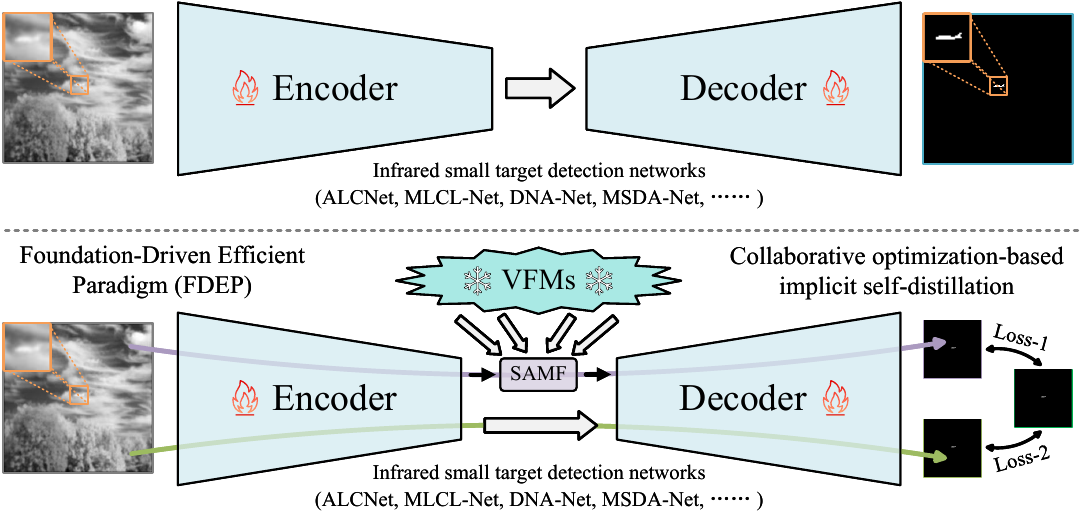}
   \caption{Comparison of different SIRST detection paradigms. \textbf{Top:} Traditional encoder-decoder-based architecture performs feature encoding and decoding in a sequential manner. \textbf{Bottom:} Our FDEP framework consists of a main branch (\textcolor{darkpurple}{purple arrow}) and a lightweight branch (\textcolor{darkgreen}{green arrow}). The main branch integrates frozen semantic representations from VFMs. The lightweight branch maintains the original structure. During inference, only the lightweight branch is used for efficiency.}
   \label{fig:figs-different-paradigms}
   \vspace{-3pt}
\end{figure}

Early SIRST studies focused on model-driven methods~\citep{qin2019infrared,arce1987theoretical,tomasi1998bilateral,li2010robust,chen2014novel,qi2012robust,chen2013local,wei2016multiscale,deng2016small,gao2013infrared,he2015small,zhang2018infrared,dai2017non}, which  explicitly model infrared imaging and background priors but are often limited by handcrafted assumptions. With the rise of deep learning, data-driven methods have shown superior adaptability and robustness by automatically learning high-dimensional mappings from data~\cite{yu2025relational,he2016deep,vaswani2017attention}. Recently, deep learning-based methods~\citep{dai2021asymmetric,zhang2023attention,li2022dense,wu2022uiu,liu2024infrared,wu2023mtu,yang2024pbt,yuan2024sctransnet,zhang2022isnet,zhao2023gradient,lin2023learning,li2025multi,zhang2024irsam,zhang2025saist,huang2025text,li2025ilnet,zhao2025multi,yu2022infrared,yu2022pay} further enhance weak-target perception by incorporating task-specific cues, such as local contrast~\citep{yu2022infrared,yu2022pay}, text~\citep{zhang2025saist,huang2025text}, and shape priors~\citep{zhang2022isnet,zhao2023gradient}. Nevertheless, three limitations remain: \textit{1. Insufficient visual global context and semantic modeling.} Existing methods focus on architectural modifications, local clue amplification, and text incorporation, but their ability to characterize visual global structural context and high-level semantic relations is relatively limited, resulting in unstable performance in complex backgrounds. \textit{2. Limited performance-efficiency trade-off.} In pursuit of higher accuracy, some methods introduce substantial parameters and computational overhead, increasing latency and limiting their applicability to real-time or resource-constrained scenarios. \textit{3. Fragmented and insufficient evaluation system.} Existing single metrics are insufficient to comprehensively assess detection capabilities, and the current evaluation system is highly sensitive to threshold settings.

To address these challenges, we rethink the SIRST detection paradigm. As shown in Fig.~\ref{fig:figs-different-paradigms}, this work first systematically introduces frozen representations from visual foundation models (VFMs) into SIRST detection, and proposes a Foundation-Driven Efficient Paradigm (FDEP). It provides a general and efficient framework compatible with diverse VFMs and SIRST networks, bridging the modality gap between natural and infrared imagery, and transferring VFM-derived priors to lightweight detectors without additional inference overhead. Specifically, considering the semantic and scale discrepancies between the high-level representations from VFMs and the task-specific features of SIRST, we design a Semantic Alignment and Modulated Fusion (SAMF) module, which achieves deep fusion between global semantic priors and task-specific representations through semantic alignment and dual-path modulation. Meanwhile, to avoid the inference time overhead introduced by VFMs and to mitigate the optimization instability and insufficient knowledge transfer of explicit distillation under extreme class imbalance, we further propose a Collaborative Optimization-based Implicit Self-Distillation (CO-ISD) strategy. By sharing encoder-decoder parameters and performing synchronous backpropagation within a unified optimization space, CO-ISD enables implicit semantic transfer between the main and lightweight branches during training, achieving stable performance gains while preserving the original detector architecture and computational cost at inference. In addition, to overcome the fragmentation and insufficiency of existing evaluation systems, we construct a Holistic SIRST Evaluation ($HSE$) metric, which performs multi-threshold integral assessment at both pixel-level confidence prediction and target-level robust detection perspectives, providing a more stable and comprehensive metric for fair model comparison and analysis. The main contributions of this work are summarized as follows:
\begin{itemize}[leftmargin=*]
    \item We systematically introduce frozen VFM representations into SIRST detection and propose FDEP, a general and efficient framework compatible with diverse VFMs and SIRST networks, which improves detection accuracy without additional VFM-related inference overhead.
    \item To bridge the semantic gap, we propose a SAMF module that achieves deep fusion of global semantic priors and task-specific representations through semantic alignment and dual-path modulation.
    \item A CO-ISD strategy is proposed to enable implicit semantic transfer through shared encoder-decoder and synchronous backpropagation, improving performance while enabling lightweight deployment.
    \item To overcome the limitations of existing evaluation systems, we construct the $HSE$ metric that performs multi-threshold integral assessment from both pixel-level and target-level perspectives.
\end{itemize}

\section{Related Work}
\label{sec:related_work}
\textbf{\textit{Non-deep learning-based SIRST detection.}} These methods rely on handcrafted features and priors, including background suppression, human visual system, and image structure-based methods. Background suppression-based methods~\cite{qin2019infrared,arce1987theoretical,tomasi1998bilateral,li2010robust,chen2014novel} enhance target responses by reducing background energy, but often degrade in complex scenes. Human visual system-based methods~\cite{qi2012robust,chen2013local,wei2016multiscale,deng2016small} exploit visual attention and contrast mechanisms, but often produce false alarms in noisy backgrounds. Image structure-based methods~\cite{gao2013infrared,he2015small,zhang2018infrared,dai2017non} formulate detection as low-rank sparse decomposition, but are computationally expensive and sensitive to complex textures. 

\textbf{\textit{Deep learning-based SIRST detection.}} Compared with non-deep learning methods, deep learning-based methods show stronger robustness and generalization~\cite{zhao2025towards,li2024icpr}. Existing methods can be broadly grouped into network structure optimization, contrast modeling, edge/shape enhancement, and text incorporation. Network structure optimization methods  mainly build on U-Net variants to improve feature extraction and semantic modeling, including CNN-based~\cite{dai2021asymmetric,zhang2023attention,li2022dense,wu2022uiu,liu2024infrared} and Transformer-hybrid networks~\cite{wu2023mtu,yang2024pbt,yuan2024sctransnet}. Contrast modeling methods~\cite{dai2021attentional,yu2022infrared,yu2022pay} inherit the idea of local contrast to enhance target salience and foreground-background separability. To better capture fine-grained edges, recent works~\cite{zhang2022isnet,zhao2023gradient,lin2023learning,li2025multi,li2025ilnet,zhao2025multi} explore edge/shape enhancement, effectively improving boundary integrity through gradient, shape, and directional cues. Furthermore, some works~\cite{zhang2025saist,huang2025text} introduce textual information to enhance detection performance. Despite their progress, existing methods focus on individual network optimization, with limited exploration of unified detection paradigms. More importantly, the potential of large-scale VFMs for SIRST detection has not yet been fully explored. Therefore, we systematically introduce VFMs into SIRST detection and propose a general and efficient framework that can be adapted to diverse VFMs and SIRST networks.

\begin{figure*}[t]
  %\captionsetup{skip=4pt}
  \centering
   \includegraphics[width=\linewidth]{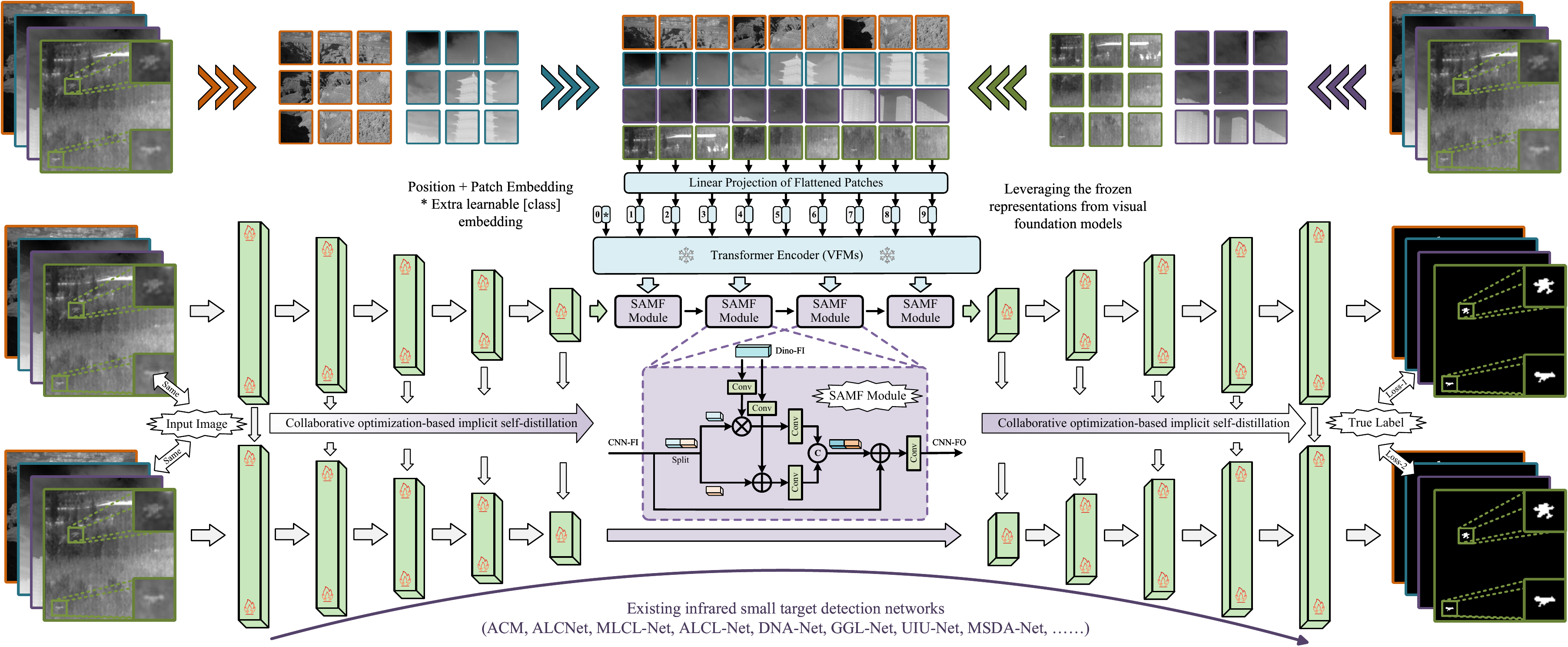}
   \caption{Overview of the FDEP Framework.}
   \label{fig:figs-fdep-framework}
   \vspace{-10pt}
\end{figure*}

\section{Methods}
\label{sec:methods}
\subsection{FDEP Framework}
To further improve the detection performance of existing SIRST networks, we propose the FDEP framework. As shown in Fig.~\ref{fig:figs-fdep-framework}, FDEP consists of a frozen VFM representation branch, a main branch, and a lightweight branch. The frozen VFM branch extracts global semantic priors from the input infrared image, activating the universal visual knowledge learned from large-scale natural images. The main branch uses the encoder of an existing SIRST network to extract task-specific high-level semantic features, and employs the proposed SAMF module to deeply align and fuse them with VFM-derived global semantic priors. The main branch features after the SAMF module not only have stronger discriminative abilities, but also significantly improve robust detection capabilities in complex backgrounds. For the lightweight branch, we retain the structure of the original SIRST detection network to ensure efficient inference. During training optimization, the proposed CO-ISD strategy is used. By sharing the encoder-decoder structure and performing synchronous backpropagation, the main branch dynamically guides the lightweight branch to learn its high-level semantic representation, thereby achieving implicit knowledge transfer and efficient alignment.

\subsection{SAMF Module}
Although VFMs (such as DINOv2~\cite{oquab2023dinov2}) can learn powerful global representations from large-scale natural images, their feature distributions remain misaligned with SIRST-specific semantics. To bridge the semantic gap, we design a SAMF module that dynamically aligns and fuses VFM-derived global semantic priors with task-specific discriminative features at the feature level. From Fig.~\ref{fig:figs-fdep-framework}, SAMF consists of three steps: semantic alignment, dual-path modulation, and residual refinement. Taking MSDA-Net equipped with the FDEP framework as an example, the process is as follows:

For semantic alignment, given multi-level features $F_{{\rm{DINO}}}^{(i)} \in {\mathbb{R}^{{C_d} \times {H_i} \times {W_i}}}, i \in \{ 6,12,18,24\}$ from the frozen DINOv2 (ViT-L/14), we first project them to the spatial scale of the backbone encoder’s top-level feature $F^{(t)} \in \mathbb{R}^{64 \times H \times W}$ using a $1\times1$ convolution followed by bilinear upsampling:
\begin{gather}
{\tilde P^{(i)}} = {\rm{Up}}({\rm{Con}}{{\rm{v}}_{1 \times 1}}(F_{{\rm{DINO}}}^{(i)})),
\end{gather}
where ${\tilde P^{(i)}} \in {\mathbb{R}^{32 \times H \times W}}$ denotes the aligned semantic prior. Subsequently, two modulation maps are generated through two independent 1×1 convolutional branches:
\begin{gather}
M_{{\rm{mul}}}^{(i)} = {\rm{Sigmoid}}({\rm{Con}}{{\rm{v}}_{1 \times 1}}({\tilde P^{(i)}})), \\
M_{{\rm{add}}}^{(i)} = {\rm{ReLU}}({\rm{BN}}({\rm{Con}}{{\rm{v}}_{1 \times 1}}({\tilde P^{(i)}}))),
\end{gather}
where ${M_{mul}}$ is a multiplicative gating map. ${M_{add}}$ is an additive semantic map.

For dual-path modulation, we divide the backbone features ${F^{(t)}}$ into two equal parts $[F_A^{(t)},{\mkern 1mu} F_B^{(t)}]$ along the channel dimension. The $F_A^{(t)}$ performs multiplicative modulation to suppress noise responses, while the $F_B^{(t)}$ performs additive modulation to inject semantic information:
\begin{gather}
B_{{\rm{mul}}}^{(i)} = {\rm{Con}}{{\rm{v}}_{3 \times 3}}(F_A^{(t)} \odot M_{{\rm{mul}}}^{(i)}), \\
B_{{\rm{add}}}^{(i)} = {\rm{Con}}{{\rm{v}}_{1 \times 1}}(F_B^{(t)} + M_{{\rm{add}}}^{(i)}),
\end{gather}
where $\odot$ denotes element-wise multiplication.

For residual refinement, the dual-path modulated features are channel-wise concatenated, then residually superimposed with the original features, and finally integrated using a 3×3 convolution:
\begin{gather}
{F^{(t + 1)}} = {\rm{Con}}{{\rm{v}}_{3 \times 3}}(([B_{{\rm{mul}}}^{(i)},{\mkern 1mu} B_{{\rm{add}}}^{(i)}]) + {F^{(t)}}),
\end{gather}
where $F^{(t + 1)}$ denotes the fused feature after the $(t+1)_{th}$ SAMF module. The top-level features of the backbone encoder are progressively fused with the features of the four DINO deep layers $(i \in \{ 6,12,18,24\} )$ to obtain the final semantically aligned fused features $F^4 \in {\mathbb{R}^{64 \times H \times W}}$.

Overall, guided by the global semantic priors from frozen VFMs, SAMF aligns and fuses task-specific and global semantic features through a three-stage mechanism. In addition, by progressively injecting multi-level semantic priors and performing decoupled feature fusion, stacked SAMF modules transfer the global prior knowledge to the task-specific detector, yielding semantically consistent, noise-resistant, and target-sensitive representations that significantly improve detection performance.

\renewcommand{\algorithmicrequire}{\textbf{Input:}}
\renewcommand{\algorithmicensure}{\textbf{Output:}}
\renewcommand{\thealgorithm}{1}
\begin{algorithm}[t]
\caption{Computation of Holistic SIRST Evaluation (\textit{HSE})}
\label{alg:hse}
\begin{algorithmic}[1]
\Require Predicted probability maps $\{P_i\}_{i=1}^{N}$; Ground truth masks $\{G_i\}_{i=1}^{N}$;
Threshold set $\mathcal{T} = \{t_j\}_{j=1}^{M}$; Centroid matching tolerance $\tau$. 
\Ensure Pixel-level \textit{HSE}-\textit{P}, Target-level \textit{HSE}-\textit{T}, Holistic \textit{HSE}.
\State\textbf{Initialize:} Give initial value to $M$, $\tau$, and Precision–Recall lists $\mathcal{P} \gets \emptyset$, $\mathcal{R} \gets \emptyset$
\State \textbf{Step HSE-P-1}: Flatten all $\{P_i\}_{i=1}^{N}$ and corresponding $\{G_i\}_{i=1}^{N}$:

\vspace{2pt}
\centerline{$\{(p_k, g_k)\}_{k=1}^{K} \leftarrow \text{Flatten}(\{(P_i, G_i)\})$}
\vspace{2pt}

\State \textbf{Step HSE-P-2}: Compute the Precision-Recall curve and integrate it to obtain the $\text{\textit{HSE}-\textit{P}}$ according to Eq.(\ref{eq:eq_hse_p}).

\State \textbf{Step HSE-T-1}: Iterate over thresholds to compute successive $(\mathcal{P},\, \mathcal{R})$ pairs:

\vspace{2pt}
\For{each threshold $t \in \mathcal{T}$}
    \State $\widehat{P}_i(t) \gets \mathbb{I}(P_i > t)$  \Comment{Perform binarization to generate the binary masks}
    \State $(\widehat{\mathcal{C}}_{\text{p}}(t),\, \mathcal{C}_{\text{g}}) \gets \text{ConnA} 
    (\widehat{P}_i(t),\, G_i)$  \Comment{Obtain connected areas from the masks}
    \State $\mathcal{M}(t) \gets \text{Match}(\widehat{\mathcal{C}}_{\text{p}}(t),\, \mathcal{C}_{\text{g}},\, \tau)$ \Comment{Conduct centroid matching}
    
    \State $(N_{\text{match}}(t),\, N_{\text{p}}(t),\, N_{\text{g}}) \gets |\mathcal{M}(t)|,\, |\widehat{\mathcal{C}}_{\text{p}}(t)|,\, |\mathcal{C}_{\text{g}}|$ \Comment{Statistical count}
    
    \If{$N_{\text{p}}(t) = 0$}     
        \State $\text{Precision}(t) \gets 0$   \Comment{Set Precision to 0 to avoid division-by-zero}
    \Else
        \State $\text{Precision}(t) \gets N_{\text{match}}(t) / N_{\text{p}}(t)$ \Comment{Compute the Precision at threshold $t$}
    \EndIf
    \State $\text{Recall}(t) \gets N_{\text{match}}(t) / N_{\text{g}}$   \Comment{Compute the Recall at threshold $t$}
    \State $(\mathcal{P},\,\! \mathcal{R}) \!\gets\! (\mathcal{P},\,\! \mathcal{R}) \!\cup\! \{(\text{Precision}(t),\,\! \text{Recall}(t))\}$ \Comment{Append to the list}
\EndFor
\vspace{2pt}

\State \textbf{Step HSE-T-2}: Integrate the Precision-Recall pairs $(\mathcal{P},\, \mathcal{R})$ across all thresholds to obtain the \textit{HSE}-\textit{T} according to Eq.(\ref{eq:eq_hse_t}).

\State \textbf{Step HSE-FINAL}: Fuse pixel-level \textit{HSE}-\textit{P} and target-level \textit{HSE}-\textit{T} to obtain the final \textit{HSE} according to Eq.(\ref{eq:eq_hse}).

\State \textbf{Return:} $\text{\textit{HSE}-\textit{P}}$, $\text{\textit{HSE}-\textit{T}}$, and $\text{\textit{HSE}}$.
\end{algorithmic}
\end{algorithm}

\subsection{CO-ISD Strategy}
To avoid the inference time overhead introduced by VFMs, we propose the CO-ISD strategy. From Fig.~\ref{fig:figs-fdep-framework}, the CO-ISD strategy consists of a main branch and a lightweight branch that are jointly optimized during training. The main branch integrates frozen VFM representations to learn globally contextual and highly discriminative features, while the lightweight branch retains the original SIRST network structure to enable efficient inference during deployment. Notably, the two branches share the same encoder-decoder parameters and compute their respective detection losses simultaneously during training. This strategy achieves implicit modeling of semantic consistency and knowledge transfer through dynamic collaborative optimization.

Specifically, the optimization objectives for the main and lightweight branches are formulated as:
\begin{gather}
{{\cal L}_{{\rm{main}}}} = {{\cal L}_{EEDM}}({Y_{{\rm{main}}}},{Y_{{\rm{gt}}}}), \\
{{\cal L}_{light}} = {{\cal L}_{EEDM}}({Y_{light}},{Y_{{\rm{gt}}}}),
\end{gather}
where ${Y_{{\rm{main}}}}$ and ${Y_{light}}$ denote the outputs of the main and lightweight branches, ${Y_{gt}}$ denotes the ground truth, and ${{\cal L}_{EEDM}}$ is the edge-enhanced difficulty-mining loss~\cite{yu2025easy,zhao2025towards}. During training, the two branches share parameters for synchronized backpropagation, and the overall objective is:
\begin{gather}
{{\cal L}_{CO \text{-} ISD}} ={{\cal L}_{main}} + \alpha  \cdot {{\cal L}_{light}}
\end{gather}
where ${{\cal L}_{CO \text{-} ISD}}$ denotes the total loss. $\alpha$ is set to 1.

Unlike explicit distillation, CO-ISD introduces no extra feature- or logit-matching loss. Instead, VFM-enhanced gradients from the main branch implicitly regularize the shared encoder-decoder parameters through synchronous optimization, enabling the lightweight branch to inherit semantic priors. After training, the VFM branch and SAMF modules are removed, and only the lightweight branch is retained for standalone inference without additional VFM-related overhead.

% Unlike explicit distillation, CO-ISD introduces no additional feature- or output-matching loss between separate teacher and student networks. Instead, the main and lightweight branches share encoder-decoder parameters and are supervised by the same ground-truth labels using the same detection loss. This avoids explicit cross-network alignment and reduces potential interference from additional distillation objectives under severe class imbalance. Gradients from both branches jointly update the shared parameters, allowing the VFM-enhanced main branch to influence the network retained for deployment. This shared, task-consistent optimization provides a mechanism for implicit knowledge transfer. After training, the VFM branch and SAMF modules are removed, and only the lightweight branch is retained for standalone inference without additional VFM-related overhead.

\begin{table*}[t]
\caption{Comparison of different methods on the SIRST3 dataset. \textit{``+ FDEP w/o CO-ISD''} denotes the FDEP without using CO-ISD. Best and second-best results are highlighted in \textcolor{bestbg_deep}{red} and \textcolor{secondbg_deep}{blue}.}
\vspace{-6pt}
\label{tab:tab01}
\setlength{\tabcolsep}{1.2mm}
\renewcommand{\arraystretch}{1.1}
\resizebox{\linewidth}{!}{
\begin{tabular}{c|c|cccc|ccc|cc}
\hline
                                                &                                           & \multicolumn{4}{c|}{Conventional Accuracy Metrics}                                                                                                                                                                                                                                                                           & \multicolumn{3}{c|}{HSE Series Accuracy Metrics}                                                                                                                                                                                                                 & \multicolumn{2}{c}{Computational Cost}                                                                                     \\ \cline{3-11} 
\multirow{-2}{*}{Methods}                       & \multirow{-2}{*}{Description}             & \multicolumn{1}{c|}{IoU ↑}                                                         & \multicolumn{1}{c|}{nIoU ↑}                                                        & \multicolumn{1}{c|}{$\mathrm{P_d}$ ↑}                                                          & $\mathrm{F_a}$ ↓                                                          & \multicolumn{1}{c|}{HSE-P ↑}                                                               & \multicolumn{1}{c|}{HSE-T ↑}                                                                & HSE ↑                                                                 & \multicolumn{1}{c|}{Param. (M)}                             & Time (s)                                                      \\ \hline
                                                & Original                                  & \multicolumn{1}{c|}{62.76}                                                         & \multicolumn{1}{c|}{61.25}                                                         & \multicolumn{1}{c|}{94.62}                                                         & {\second{18.78}}                                                   & \multicolumn{1}{c|}{82.41}                                                                 & \multicolumn{1}{c|}{82.58}                                                                  & 68.05                                                                 & \multicolumn{1}{c|}{{\textbf{0.40}}}  & {\textbf{0.022}}                         \\ \cline{2-11} 
                                                & \cellcolor{fdepbg}+ FDEP w/o CO-ISD & \multicolumn{1}{c|}{{\best{70.43}}} & \multicolumn{1}{c|}{{\best{69.73}}} & \multicolumn{1}{c|}{{\best{96.48}}} & \cellcolor{fdepbg}19.70                                 & \multicolumn{1}{c|}{{\best{90.02 ($\boldsymbol{\Uparrow}$ 7.61)}}} & \multicolumn{1}{c|}{{\best{86.66 ($\boldsymbol{\Uparrow}$ 
 4.08)}}}  & {\best{78.01 ($\boldsymbol{\Uparrow}$ 9.96)}} & \multicolumn{1}{c|}{\cellcolor{fdepbg}0.85}    & \cellcolor{fdepbg}0.082                          \\ \cline{2-11} 
\multirow{-3}{*}{ACM~\cite{dai2021asymmetric}}      & \cellcolor{fdepbg}+ FDEP Framework  & \multicolumn{1}{c|}{\second{68.56}}                           & \multicolumn{1}{c|}{\second{68.01}}                           & \multicolumn{1}{c|}{\second{95.42}}                           & {\best{13.92}} & \multicolumn{1}{c|}{\second{89.17 ($\Uparrow$ 6.76)}}                           & \multicolumn{1}{c|}{\second{85.02 ($\Uparrow$ 2.44)}}                            & \second{75.81 ($\Uparrow$ 7.76)}                           & \multicolumn{1}{c|}{\cellcolor{fdepbg}0.85}    & {\cellcolor{fdepbg}\textbf{0.022}} \\ \hline
                                                & Original                                  & \multicolumn{1}{c|}{68.30}                                                         & \multicolumn{1}{c|}{69.55}                                                         & \multicolumn{1}{c|}{95.28}                                                         & 34.07                                                         & \multicolumn{1}{c|}{87.56}                                                                 & \multicolumn{1}{c|}{84.74}                                                                  & 74.20                                                                 & \multicolumn{1}{c|}{{\textbf{0.43}}}  & {\textbf{0.021}}                         \\ \cline{2-11} 
                                                & \cellcolor{fdepbg}+ FDEP w/o CO-ISD & \multicolumn{1}{c|}{{\best{73.17}}} & \multicolumn{1}{c|}{{\best{73.23}}} & \multicolumn{1}{c|}{{\best{97.14}}} & \second{24.95}                           & \multicolumn{1}{c|}{{\best{91.73 ($\boldsymbol{\Uparrow}$ 4.17)}}} & \multicolumn{1}{c|}{{\best{88.67 ($\boldsymbol{\Uparrow}$ 3.93)}}}  & {\best{81.34 ($\boldsymbol{\Uparrow}$ 7.14)}} & \multicolumn{1}{c|}{\cellcolor{fdepbg}0.88}    & \cellcolor{fdepbg}0.082                           \\ \cline{2-11} 
\multirow{-3}{*}{ALCNet~\cite{dai2021attentional}}   & \cellcolor{fdepbg}+ FDEP Framework  & \multicolumn{1}{c|}{\second{70.57}}                           & \multicolumn{1}{c|}{\second{72.88}}                           & \multicolumn{1}{c|}{\second{95.61}}                           & {\best{22.05}} & \multicolumn{1}{c|}{\second{89.75 ($\Uparrow$ 2.19)}}                           & \multicolumn{1}{c|}{\second{87.25 ($\Uparrow$ 2.51)}}                            & \second{78.31 ($\Uparrow$ 4.11)}                           & \multicolumn{1}{c|}{\cellcolor{fdepbg}0.88}    & {\cellcolor{fdepbg}\textbf{0.021}} \\ \hline
                                                & Original                                  & \multicolumn{1}{c|}{79.83}                                                         & \multicolumn{1}{c|}{82.88}                                                         & \multicolumn{1}{c|}{\second{96.35}}                                                   & 19.02                                                         & \multicolumn{1}{c|}{\second{94.37}}                                                           & \multicolumn{1}{c|}{91.98}                                                                  & 86.80                                                                 & \multicolumn{1}{c|}{{\textbf{0.66}}}  & {\textbf{0.025}}                         \\ \cline{2-11} 
                                                & \cellcolor{fdepbg}+ FDEP w/o CO-ISD & \multicolumn{1}{c|}{{\best{84.16}}} & \multicolumn{1}{c|}{{\best{85.08}}} & \multicolumn{1}{c|}{{\best{97.54}}} & {\best{8.59}}  & \multicolumn{1}{c|}{{\best{96.63 ($\boldsymbol{\Uparrow}$ 2.26)}}} & \multicolumn{1}{c|}{{\best{93.57 ($\boldsymbol{\Uparrow}$ 1.59)}}}  & {\best{90.43 ($\boldsymbol{\Uparrow}$ 3.63)}} & \multicolumn{1}{c|}{\cellcolor{fdepbg}1.11}    & \cellcolor{fdepbg}0.087                           \\ \cline{2-11} 
\multirow{-3}{*}{MLCL-Net~\cite{yu2022infrared}}  & \cellcolor{fdepbg}+ FDEP Framework  & \multicolumn{1}{c|}{\second{80.24}}                           & \multicolumn{1}{c|}{\second{83.46}}                           & \multicolumn{1}{c|}{\cellcolor{fdepbg}95.42}                                 & \second{12.25}                           & \multicolumn{1}{c|}{\cellcolor{fdepbg}94.27 ($\Downarrow$ 0.10)}                                 & \multicolumn{1}{c|}{\second{92.50 ($\Uparrow$ 0.52)}}                            & \second{87.20 ($\Uparrow$ 0.40)}                           & \multicolumn{1}{c|}{\cellcolor{fdepbg}1.11}    & {\cellcolor{fdepbg}\textbf{0.025}} \\ \hline
                                                & Original                                  & \multicolumn{1}{c|}{80.94}                                                         & \multicolumn{1}{c|}{83.39}                                                         & \multicolumn{1}{c|}{\second{96.74}}                                                   & \second{11.22}                                                   & \multicolumn{1}{c|}{95.09}                                                                 & \multicolumn{1}{c|}{91.23}                                                                  & 86.75                                                                 & \multicolumn{1}{c|}{{\textbf{5.67}}}  & {\textbf{0.024}}                         \\ \cline{2-11} 
                                                & \cellcolor{fdepbg}+ FDEP w/o CO-ISD & \multicolumn{1}{c|}{{\best{83.06}}} & \multicolumn{1}{c|}{{\best{84.16}}} & \multicolumn{1}{c|}{{\best{97.21}}} & {\best{8.49}}  & \multicolumn{1}{c|}{{\best{96.31 ($\boldsymbol{\Uparrow}$ 1.22)}}} & \multicolumn{1}{c|}{{\best{92.30 ($\boldsymbol{\Uparrow}$ 1.07)}}}  & {\best{88.90 ($\boldsymbol{\Uparrow}$ 2.15)}} & \multicolumn{1}{c|}{\cellcolor{fdepbg}9.74}    & \cellcolor{fdepbg}0.086                           \\ \cline{2-11} 
\multirow{-3}{*}{ALCL-Net~\cite{yu2022pay}} & \cellcolor{fdepbg}+ FDEP Framework  & \multicolumn{1}{c|}{\second{81.83}}                           & \multicolumn{1}{c|}{\second{83.95}}                           & \multicolumn{1}{c|}{\cellcolor{fdepbg}96.41}                                 & \cellcolor{fdepbg}16.09                                 & \multicolumn{1}{c|}{\second{95.74 ($\Uparrow$ 0.65)}}                           & \multicolumn{1}{c|}{\second{91.29 ($\Uparrow$ 0.06)}}                            & \second{87.40 ($\Uparrow$ 0.65)}                           & \multicolumn{1}{c|}{\cellcolor{fdepbg}9.74}    & {\cellcolor{fdepbg}\textbf{0.024}} \\ \hline
                                                & Original                                  & \multicolumn{1}{c|}{\second{82.59}}                                                   & \multicolumn{1}{c|}{\second{86.08}}                                                   & \multicolumn{1}{c|}{96.68}                                                         & \second{9.29}                                                    & \multicolumn{1}{c|}{95.38}                                                                 & \multicolumn{1}{c|}{93.41}                                                                  & 89.10                                                                 & \multicolumn{1}{c|}{{\textbf{4.70}}}  & {\textbf{0.095}}                         \\ \cline{2-11} 
                                                & \cellcolor{fdepbg}+ FDEP w/o CO-ISD & \multicolumn{1}{c|}{{\best{83.06}}} & \multicolumn{1}{c|}{{\best{86.09}}} & \multicolumn{1}{c|}{{\best{98.14}}} & {\best{8.90}}  & \multicolumn{1}{c|}{{\best{96.33 ($\boldsymbol{\Uparrow}$ 0.95)}}} & \multicolumn{1}{c|}{{\best{94.99 ($\boldsymbol{\Uparrow}$ 1.58)}}} & {\best{91.50 ($\boldsymbol{\Uparrow}$ 2.40)}} & \multicolumn{1}{c|}{\cellcolor{fdepbg}8.77}    & \cellcolor{fdepbg}0.160                          \\ \cline{2-11} 
\multirow{-3}{*}{DNA-Net~\cite{li2022dense}}   & \cellcolor{fdepbg}+ FDEP Framework  & \multicolumn{1}{c|}{\cellcolor{fdepbg}82.54}                                 & \multicolumn{1}{c|}{\cellcolor{fdepbg}86.06}                                 & \multicolumn{1}{c|}{\second{97.54}}                           & \cellcolor{fdepbg}11.93                                 & \multicolumn{1}{c|}{\second{95.46 ($\Uparrow$ 0.08)}}                           & \multicolumn{1}{c|}{\second{93.89 ($\Uparrow$ 0.48)}}                            & \second{89.62 ($\Uparrow$ 0.52)}                           & \multicolumn{1}{c|}{\cellcolor{fdepbg}8.77}    & {\cellcolor{fdepbg}\textbf{0.095}} \\ \hline
                                                & Original                                  & \multicolumn{1}{c|}{81.69}                                                         & \multicolumn{1}{c|}{84.74}                                                         & \multicolumn{1}{c|}{97.21}                                                         & {\best{7.78}}                          & \multicolumn{1}{c|}{95.24}                                                                 & \multicolumn{1}{c|}{92.39}                                                                  & 87.99                                                                 & \multicolumn{1}{c|}{{\textbf{8.99}}}  & {\textbf{0.036}}                         \\ \cline{2-11} 
                                                & \cellcolor{fdepbg}+ FDEP w/o CO-ISD & \multicolumn{1}{c|}{{\best{84.35}}} & \multicolumn{1}{c|}{\second{85.62}}                           & \multicolumn{1}{c|}{{\best{97.54}}} & \cellcolor{fdepbg}13.90                                 & \multicolumn{1}{c|}{{\best{96.78 ($\boldsymbol{\Uparrow}$ 1.54)}}} & \multicolumn{1}{c|}{{\best{94.58 ($\boldsymbol{\Uparrow}$ 2.19)}}}  & {\best{91.53 ($\boldsymbol{\Uparrow}$ 3.54)}} & \multicolumn{1}{c|}{\cellcolor{fdepbg}13.06}   & \cellcolor{fdepbg}0.100                         \\ \cline{2-11} 
\multirow{-3}{*}{GGL-Net~\cite{zhao2023gradient}}  & \cellcolor{fdepbg}+ FDEP Framework  & \multicolumn{1}{c|}{\second{83.59}}                           & \multicolumn{1}{c|}{{\best{86.29}}} & \multicolumn{1}{c|}{\second{97.34}}                           & \second{11.46}                           & \multicolumn{1}{c|}{\second{96.48 ($\Uparrow$
 1.24)}}                           & \multicolumn{1}{c|}{\second{94.00 ($\Uparrow$ 1.61)}}                            & \second{90.69 ($\Uparrow$ 2.70)}                           & \multicolumn{1}{c|}{\cellcolor{fdepbg}13.06}   & {\cellcolor{fdepbg}\textbf{0.036}} \\ \hline
                                                & Original                                  & \multicolumn{1}{c|}{81.99}                                                         & \multicolumn{1}{c|}{83.73}                                                         & \multicolumn{1}{c|}{96.54}                                                         & \second{12.15}                                                   & \multicolumn{1}{c|}{96.00}                                                                 & \multicolumn{1}{c|}{91.69}                                                                  & 88.02                                                                 & \multicolumn{1}{c|}{{\textbf{50.54}}} & {\textbf{0.055}}                         \\ \cline{2-11} 
                                                & \cellcolor{fdepbg}+ FDEP w/o CO-ISD & \multicolumn{1}{c|}{{\best{84.34}}} & \multicolumn{1}{c|}{{\best{85.58}}} & \multicolumn{1}{c|}{{\best{97.41}}} & {\best{5.35}}  & \multicolumn{1}{c|}{{\best{96.93 ($\boldsymbol{\Uparrow}$ 0.93)}}} & \multicolumn{1}{c|}{{\best{93.30 ($\boldsymbol{\Uparrow}$ 1.61)}}}  & {\best{90.44 ($\boldsymbol{\Uparrow}$ 2.42)}} & \multicolumn{1}{c|}{\cellcolor{fdepbg}64.71}   & \cellcolor{fdepbg}0.120                           \\ \cline{2-11} 
\multirow{-3}{*}{UIU-Net~\cite{wu2022uiu}}   & \cellcolor{fdepbg}+ FDEP Framework  & \multicolumn{1}{c|}{\second{83.12}}                           & \multicolumn{1}{c|}{\second{85.05}}                           & \multicolumn{1}{c|}{\second{97.08}}                           & \cellcolor{fdepbg}13.31                                 & \multicolumn{1}{c|}{\second{96.57 ($\Uparrow$ 0.57)}}                           & \multicolumn{1}{c|}{\second{93.22 ($\Uparrow$ 1.53)}}                            & \second{90.01 ($\Uparrow$ 1.99)}                           & \multicolumn{1}{c|}{\cellcolor{fdepbg}64.71}   & {\cellcolor{fdepbg}\textbf{0.055}} \\ \hline
                                                & Original                                  & \multicolumn{1}{c|}{82.86}                                                         & \multicolumn{1}{c|}{85.46}                                                         & \multicolumn{1}{c|}{97.01}                                                         & \second{12.53}                                                   & \multicolumn{1}{c|}{96.43}                                                                 & \multicolumn{1}{c|}{93.04}                                                                  & 89.72                                                                 & \multicolumn{1}{c|}{{\textbf{4.79}}}  & {\textbf{0.035}}                         \\ \cline{2-11} 
                                                & \cellcolor{fdepbg}+ FDEP w/o CO-ISD & \multicolumn{1}{c|}{{\best{84.71}}} & \multicolumn{1}{c|}{{\best{86.57}}} & \multicolumn{1}{c|}{{\best{98.14}}} & {\best{11.95}} & \multicolumn{1}{c|}{{\best{97.25 ($\boldsymbol{\Uparrow}$ 0.82)}}} & \multicolumn{1}{c|}{{\best{94.32 ($\boldsymbol{\Uparrow}$ 1.28)}}}  & {\best{91.73 ($\boldsymbol{\Uparrow}$ 2.01)}} & \multicolumn{1}{c|}{\cellcolor{fdepbg}5.24}    & \cellcolor{fdepbg}0.100                           \\ \cline{2-11} 
\multirow{-3}{*}{MSDA-Net~\cite{zhao2025multi}} & \cellcolor{fdepbg}+ FDEP Framework  & \multicolumn{1}{c|}{\second{84.05}}                           & \multicolumn{1}{c|}{\second{86.33}}                           & \multicolumn{1}{c|}{\second{97.54}}                           & \cellcolor{fdepbg}14.24                                 & \multicolumn{1}{c|}{\second{96.92 ($\Uparrow$ 0.49)}}                           & \multicolumn{1}{c|}{\second{94.21 ($\Uparrow$ 1.17)}}                            & \second{91.31 ($\Uparrow$ 1.59)}                           & \multicolumn{1}{c|}{\cellcolor{fdepbg}5.24}    & {\cellcolor{fdepbg}\textbf{0.035}} \\ \hline
\end{tabular}
}

\end{table*}

\subsection{HSE Metric}
In the SIRST task, existing evaluation protocols suffer from fragmentation and inherent limitations. Specifically, existing studies rely on a set of discrete metrics (e.g. $IoU$, $nIoU$, $P_d$, $F_a$), which fail to comprehensively reflect the quality of model-generated probability maps.  At the same time, the $ROC\text{-}AUC$ aggregates performance across thresholds but is insensitive to false positives under extreme class imbalance (please see Sec.~\ref{sec:roc_limitations}). To address these limitations, we propose a unified and comprehensive $HSE$ metric as shown in Algorithm~\ref{alg:hse}. Notably, \textit{HSE} differs fundamentally from \textit{hIoU}~\cite{pang2025rethinking}. The \textit{hIoU} adopts hierarchical discrete decisions and emphasizes target-level results by considering pixel-level IoU only for correctly detected targets. In contrast, our \textit{HSE} unifies pixel- and target-level evaluation via continuous multi-threshold integration, jointly capturing pixel-level confidence stability and target-level robustness while avoiding single-threshold bias. Tab.~\ref{tab:tabs-s05} in Appendix~\ref{sec:more Discussion} provides a detailed quantitative comparison.

The \textit{HSE} consists of two complementary sub-metrics: \textit{HSE-P} and \textit{HSE-T}. The final score:
\begin{gather} \label{eq:eq_hse}
HSE{\rm{ }} = {\rm{ }}HSE\text{-}P{\rm{ }} \times {\rm{ }}HSE\text{-}T
\end{gather}
where $HSE\text{-}P$ and $HSE\text{-}T$ evaluate pixel-level confidence quality and target-level detection robustness, respectively, and their product fusion jointly captures local precision and overall detectability.
% where $HSE$-$P$ integrates pixel-level precision-recall over dense probability maps, and $HSE$-$T$ integrates target-level centroid-matching precision-recall across thresholds. Their product encourages balanced pixel-level confidence quality and target-level detection robustness.

For $HSE\text{-}P$, predicted probability maps $\{ {p_i}\} _{i = 1}^N$ and ground truths $\{ {G_i}\} _{i = 1}^N$ are flattened and concatenated to compute the $Precision$ \text{-} $Recall$ ($PR$) curve, whose area is used as the score:
% \begin{gather}
% {\rm{HSE\text{-}P}} = \int_0^1 {{\rm{Precision}}} ({\rm{Recall}}){\mkern 1mu} d({\rm{Recall}})
% \end{gather}
% \begin{gather}
% {\rm{HSE\text{-}P}} = \int_0^1 {{\rm{P}}} ({\rm{R}}){\mkern 1mu} d({\rm{R}})
% \end{gather}
\begin{gather} \label{eq:eq_hse_p}
HSE\text{-}P = \int_0^1 P(R)\, \mathrm{d}R\text{ ,}
\end{gather}
where $P(R)$ is the $Precision$ as a function of $Recall$. Compared to $ROC\text{-}AUC$, $HSE\text{-}P$ is more sensitive to false positive pixels, offering a more reliable measure.

For $HSE\text{-}T$, we first binarize the $\{ {p_i}\} _{i = 1}^N$ under multiple threshold sets $T = \{ {t_1},{t_2}, \ldots ,{t_M}\}$, and extract the connected component sets ${\hat C_p}({t_j})$ and ${\hat C_g}$ of the predicted target and the true target. For each true target centroid ${c_g} \in {C_g}$, iterate through all unmatched predicted target centroids ${\hat c_p} \in {\hat C_p}({t_j})$. If the Euclidean distance meets the set threshold, it is regarded as a correct detection, and the predicted target is marked as “matched” to prevent duplication. Then, based on the number of successful matches, predicted targets, and true targets, $Precision$ and $Recall$ are calculated at each threshold.  Finally, $HSE\text{-}T$ is obtained by discretely integrating the \textit{Precision-Recall} pairs:
% \begin{gather}
% {\rm{HSE\text{-}T}} = \sum\limits_{j = 1}^M {{\rm{P}}} ({t_j})({\rm{R}}({t_j}) - {\rm{R}}({t_{j + 1}}))
% \end{gather}
\begin{equation}\label{eq:eq_hse_t}
HSE\text{-}T = \sum_{j=1}^{M} P(t_j)\,[R(t_j) - R(t_{j+1})],
\end{equation}
where $P(t_j)$ and $R(t_j)$ denote the $Precision$ and $Recall$ at threshold $t_j$. Unlike traditional metrics such as $P_d$ and $F_a$, $HSE\text{-}T$ more stably measures the model's detection robustness and generalization ability under different decision boundaries through discrete integration with multiple thresholds.

% In summary, the proposed $HSE$ metric upgrades the evaluation system from point-based assessment to line-based integration and ultimately to surface-based measurement in probability space, establishing a fair, comprehensive, and stable evaluation for SIRST detection.

\section{Experiments}
\label{sec:experiments}
\subsection{Datasets}
\label{sec:datasets}
We conduct experiments on four public SIRST datasets: SIRST3~\cite{yu2025easy,ying2023mapping}, NUAA-SIRST~\cite{dai2021asymmetric}, NUDT-SIRST~\cite{li2022dense} and IRSTD-1K~\cite{zhang2022isnet}. They contain 2755, 427, 1327, and 1001 samples respectively. These datasets cover a wide range of scenarios, from real-world complex scenes to synthetic multi-target environments, providing a comprehensive basis for verifying the robustness and generalization.
% The NUAA-SIRST, NUDT-SIRST, and IRSTD-1K datasets are used to verify the robustness under limited samples and complex scenarios, while the SIRST3 dataset is used to further explore the comprehensive detection performance under cross-scene, cross-target type, and multi-scale conditions.

\subsection{Implementation Details}
\label{sec:implementation_details}
\textbf{\textit{Experiment setting.}} We use the AdamW~\cite{loshchilov2017decoupled} optimizer with an initial learning rate of $1.0 \times 10^{-3}$. The batch size is 16, the number of epochs is 400, and the GPU is an RTX 4090 24 GB. During training, all input images are normalized and randomly cropped into 448 × 448 pixels image patches. Unless otherwise specified, the default VFMs used are \textit{DINOv2 (ViT-L/14)}.

\textbf{\textit{Evaluation metrics.}} To facilitate a smooth transition, in addition to $HSE$ series, we also adopt four commonly used metrics: pixel-level metrics ($IoU$ and $nIoU$) and target-level metrics ($P_d$ and $F_a$). Consistent with previous research~\cite{zhao2025towards,yu2025easy,yu2024lr,li2024icpr}, the model is invalid when $F_a > 1e^{-4}$.

\begin{table*}[t]
\caption{Comparison of different methods on the NUAA-SIRST, NUDT-SIRST, and IRSTD-1k datasets. \textit{``+ FDEP w/o CO-ISD''} denotes the FDEP framework without CO-ISD strategy.}
\vspace{-6pt}
\label{tab:tab02}
\setlength{\tabcolsep}{1mm}
\renewcommand{\arraystretch}{1.2}
\resizebox{\linewidth}{!}{
\begin{tabular}{c|c|ccccc|ccccc|ccccc}
\hline
                           &                                           & \multicolumn{5}{c|}{NUAA-SIRST (213:214)}                                                                                                                                                                                                                                                                                                                                                                         & \multicolumn{5}{c|}{NUDT-SIRST (663:664)}                                                                                                                                                                                                                                                                                                                                                                         & \multicolumn{5}{c}{IRSTD-1K (800: 201)}                                                                                                                                                                                                                                                                                                                                                                           \\
\hline
\end{tabular}
}
\vspace{-8pt}
\end{table*}

\subsection{Comparison with SOTA Methods}
We embed multiple excellent SIRST detection networks~\cite{li2022dense,yu2022infrared,yu2022pay,zhao2023gradient,dai2021attentional,zhao2025multi,dai2021asymmetric,wu2022uiu} into our framework. For fair comparison, we retrain all models under the same settings. More experimental results  are provided in Tab.~\ref{tab:tabs-s03} and Tab.~\ref{tab:tabs-s04} of Appendix~\ref{sec:compared wit more sota methods}.

\textbf{\textit{Evaluation on the SIRST3 dataset.}} From Tab.~\ref{tab:tab01}, first, when the frozen VFM is introduced (i.e., \textit{``+FDEP w/o CO-ISD''}), all networks exhibit consistent improvements, with $IoU$, $nIoU$, $P_d$, and $HSE$ increasing by 0.47 - 7.67, 0.01 - 8.48, 0.33 - 1.86, and 2.01 - 9.96, respectively. Second, compared with \textit{``+FDEP w/o CO-ISD''}, the introduction of the CO-ISD strategy further reduces inference time by 40.6\% (from 0.160 to 0.095) - 74.4\% (from 0.082 to 0.021), while maintaining comparable detection performance. For most networks (e.g., MSDA-Net), the difference in $HSE$ is less than 1.0, indicating effective representation alignment between the main and lightweight branches. Finally, the FDEP framework consistently enhances performance across diverse models without increasing inference cost, with $HSE$ improving by 0.40 - 7.76, highlighting its strong transferability and generalization. From Fig.~\ref{fig:figs-method-results} of the Appendix ~\ref{sec:more qualitative analyses}, both \textit{“+FDEP w/o CO-ISD”} and\textit{ “+FDEP Framework”} deliver comparable results, which are better than those of the original networks.

\textbf{\textit{Evaluation on three individual datasets.}} As shown in Tab.~\ref{tab:tab02}, both \textit{“+ FDEP w/o CO-ISD”} and \textit{“+ FDEP Framework”} consistently improve performance over the \textit{“Original”} across all three datasets. Specifically, \textit{“+ FDEP w/o CO-ISD”} increases the $HSE$ by 0.48 - 6.51, 0.25 - 4.38, 0.72 - 7.01 on the NUAA-SIRST, NUDT-SIRST, and IRSTD-1K datasets, respectively. Similarly, \textit{“+ FDEP Framework”} increases the $HSE$ by 0.05 - 2.31, 0.31 - 2.79, and 0.48 - 2.48, respectively. Overall, these results fully validate that our FDEP delivers stable gains under diverse challenging conditions with few-shot samples. From Figs.~\ref{fig:figs-s01}--\ref{fig:figs-s03} of the Appendix~\ref{sec:more qualitative analyses}, both \textit{“+FDEP w/o CO-ISD”} and\textit{ “+FDEP Framework”} deliver comparable results, which are better than those of the original networks.

\begin{table}[t]
\begin{minipage}[]{\linewidth}
\centering
\caption{Break-down ablation on the SIRST3 dataset.}
\setlength{\tabcolsep}{2.2mm}
\renewcommand{\arraystretch}{1.1}
\resizebox{\linewidth}{!}{
\begin{tabular}{ccllcllll}
\hline
\multicolumn{1}{c|}{} & \multicolumn{3}{c|}{Variants} & \multicolumn{5}{c}{MSDA-Net + FDEP Framework} \\ \cline{2-9}
\multicolumn{1}{c|}{\multirow{-2}{*}{Scheme}} & VFMs & \multicolumn{1}{c}{SAMF} & \multicolumn{1}{c|}{CO-ISD} &
\multicolumn{1}{c|}{HSE-P ↑} & \multicolumn{1}{c|}{HSE-T ↑} & \multicolumn{1}{c|}{HSE ↑} &
\multicolumn{1}{c|}{Param. (M) ↓} & \multicolumn{1}{c}{Time (s) ↓} \\ \hline
\multicolumn{1}{c|}{Original} & \textbf{\ding{55}} & \multicolumn{1}{c}{\textbf{\ding{55}}} & \multicolumn{1}{c|}{\textbf{\ding{55}}} &
\multicolumn{1}{c|}{96.43} & \multicolumn{1}{c|}{93.04} & \multicolumn{1}{c|}{89.72} &
\multicolumn{1}{c|}{{\textbf{4.79}}} & \multicolumn{1}{c}{{\textbf{0.037}}} \\ \hline
\multicolumn{1}{c|}{Scheme1} & \textbf{\ding{51}} & \multicolumn{1}{c}{\textbf{\ding{55}}} & \multicolumn{1}{c|}{\textbf{\ding{55}}} &
\multicolumn{1}{c|}{96.93 \gain{+0.50}} & \multicolumn{1}{c|}{94.29 \gain{+1.25}} & \multicolumn{1}{c|}{91.40 \gain{+1.68}} &
\multicolumn{1}{c|}{5.05} & \multicolumn{1}{c}{0.100} \\
\rowcolor[HTML]{D7D7D7}
\multicolumn{1}{c|}{\cellcolor[HTML]{D7D7D7}Scheme2} & \textbf{\ding{51}} &
\multicolumn{1}{c}{\cellcolor[HTML]{D7D7D7}\textbf{\ding{51}}} &
\multicolumn{1}{c|}{\cellcolor[HTML]{D7D7D7}\textbf{\ding{55}}} &
\multicolumn{1}{c|}{\cellcolor[HTML]{D7D7D7}{\textbf{97.25 \gain{+0.82}}}} &
\multicolumn{1}{c|}{\cellcolor[HTML]{D7D7D7}{\textbf{94.32 \gain{+1.28}}}} &
\multicolumn{1}{c|}{\cellcolor[HTML]{D7D7D7}{\textbf{91.73 \gain{+2.01}}}} &
\multicolumn{1}{c|}{\cellcolor[HTML]{D7D7D7}5.24} &
\multicolumn{1}{c}{\cellcolor[HTML]{D7D7D7}0.100} \\ \hline
\multicolumn{1}{c|}{Scheme3} & \textbf{\ding{51}} & \multicolumn{1}{c}{\textbf{\ding{55}}} & \multicolumn{1}{c|}{\textbf{\ding{51}}} &
\multicolumn{1}{c|}{96.76 \gain{+0.33}} & \multicolumn{1}{c|}{93.73 \gain{+0.69}} & \multicolumn{1}{c|}{90.92 \gain{+1.20}} &
\multicolumn{1}{c|}{5.05} & \multicolumn{1}{c}{{\textbf{0.037}}} \\
\rowcolor[HTML]{D7D7D7}
\multicolumn{1}{c|}{\cellcolor[HTML]{D7D7D7}Scheme4} & \cellcolor[HTML]{D7D7D7}\textbf{\ding{51}} &
\multicolumn{1}{c}{\cellcolor[HTML]{D7D7D7}\textbf{\ding{51}}} &
\multicolumn{1}{c|}{\cellcolor[HTML]{D7D7D7}\textbf{\ding{51}}} &
\multicolumn{1}{c|}{\cellcolor[HTML]{D7D7D7}{\textbf{96.92 \gain{+0.49}}}} &
\multicolumn{1}{c|}{\cellcolor[HTML]{D7D7D7}{\textbf{94.21 \gain{+1.17}}}} &
\multicolumn{1}{c|}{\cellcolor[HTML]{D7D7D7}{\textbf{91.31 \gain{+1.59}}}} &
\multicolumn{1}{c|}{\cellcolor[HTML]{D7D7D7}5.24} &
\multicolumn{1}{c}{\cellcolor[HTML]{D7D7D7}{\textbf{0.037}}} \\ \hline
\end{tabular}}
\label{tab:tab03}
% \end{table}
\end{minipage}

\vspace{5pt}

% \begin{table}[t]
\begin{minipage}[]{\linewidth}
\centering
\caption{Evaluation of the VFMs directly applied to the SIRST detection on the SIRST3 dataset.}
\label{tab:tab07}
\setlength{\tabcolsep}{1.2mm}
\renewcommand{\arraystretch}{1.1}
\resizebox{1.0\linewidth}{!}{
\begin{tabular}{c|c|cccc|ccc}
\hline
                          &                                                                          & \multicolumn{4}{c|}{Conventional Metrics}                                                                                                                                                                                                                                                            & \multicolumn{3}{c}{HSE Series Metrics}                                                                                                                                                                          \\ \cline{3-9} 
\multirow{-2}{*}{Scheme}  & \multirow{-2}{*}{\begin{tabular}[c]{@{}c@{}}VFM \\ Frozen\end{tabular}} & \multicolumn{1}{c|}{IoU ↑}                                                         & \multicolumn{1}{c|}{nIoU ↑}                                                        & \multicolumn{1}{c|}{$\mathrm{P_d}$ ↑}                                                          & $\mathrm{F_a}$ ↓                                 & \multicolumn{1}{c|}{HSE-P ↑}                                                       & \multicolumn{1}{c|}{HSE-T ↑}                                                       & HSE ↑                                 \\ \hline
VFMs + Decoder            & \textbf{\ding{51}}                                                               & \multicolumn{1}{c|}{34.15}                                                         & \multicolumn{1}{c|}{26.98}                                                         & \multicolumn{1}{c|}{55.48}                                                         & 91.11                                 & \multicolumn{1}{c|}{52.62 \gain{+44.63}}                                                         & \multicolumn{1}{c|}{41.38 \gain{+52.94}}                                                         & 21.77 \gain{+69.96}                                \\ \hline
VFMs + Decoder            & \textbf{\ding{55}}                                                               & \multicolumn{1}{c|}{33.87}                                                         & \multicolumn{1}{c|}{26.87}                                                         & \multicolumn{1}{c|}{56.81}                                                         & 80.67                                 & \multicolumn{1}{c|}{52.03 \gain{+45.22}}                                                         & \multicolumn{1}{c|}{44.27 \gain{+50.05}}                                                         & 23.03 \gain{+68.70}                                 \\ \hline
VFMs + MSDA-Net (Decoder) & \textbf{\ding{51}}                                                               & \multicolumn{1}{c|}{69.08}                                                         & \multicolumn{1}{c|}{69.40}                                                         & \multicolumn{1}{c|}{96.40}                                                         & 18.68                                 & \multicolumn{1}{c|}{90.00 \gain{+7.25}}                                                         & \multicolumn{1}{c|}{90.96 \gain{+3.36}}                                                         & 81.86 \gain{+9.87}                                \\ \hline
VFMs + MSDA-Net (Decoder) & \textbf{\ding{55}}                                                               & \multicolumn{1}{c|}{69.98}                                                         & \multicolumn{1}{c|}{70.91}                                                         & \multicolumn{1}{c|}{96.81}                                                         & 17.30                                 & \multicolumn{1}{c|}{90.31 \gain{+6.94}}                                                         & \multicolumn{1}{c|}{91.50 \gain{+2.82}}                                                         & 82.64 \gain{+9.09}                                 \\ \hline
\rowcolor[HTML]{D7D7D7} 
FDEP w/o CO-ISD           & \textbf{\ding{51}}                                                               & \multicolumn{1}{c|}{\cellcolor[HTML]{D7D7D7}{\textbf{84.71}}} & \multicolumn{1}{c|}{\cellcolor[HTML]{D7D7D7}{\textbf{86.57}}} & \multicolumn{1}{c|}{\cellcolor[HTML]{D7D7D7}{\textbf{98.14}}} & {\textbf{11.95}} & \multicolumn{1}{c|}{\cellcolor[HTML]{D7D7D7}{\textbf{97.25}}} & \multicolumn{1}{c|}{\cellcolor[HTML]{D7D7D7}{\textbf{94.32}}} & {\textbf{91.73 }} \\ \hline
\end{tabular}
}
% \end{table}
\end{minipage}

\vspace{5pt}

% \begin{table}[t]
\begin{minipage}[]{\linewidth}
\caption {Performance comparison of FDEP framework with different VFMs on the SIRST3 dataset. ``Baseline (N/A)'' denotes the original MSDA-Net without VFM integration.}
\label{tab:tab08}
\setlength{\tabcolsep}{2.2mm}
\renewcommand{\arraystretch}{1.1}
\resizebox{\linewidth}{!}{
\begin{tabular}{c|ccc|c|ccc}
\hline
                                & \multicolumn{3}{c|}{MSDA-Net + FDEP}                                                                     &                                 & \multicolumn{3}{c}{MSDA-Net + FDEP}                                                                      \\ \cline{2-4} \cline{6-8} 
\multirow{-2}{*}{VFMs (Scales)} & \multicolumn{1}{c|}{HSE-P ↑} & \multicolumn{1}{c|}{HSE-T ↑} & \cellcolor[HTML]{D7D7D7}\textbf{HSE ↑}           & \multirow{-2}{*}{VFMs (Scales)} & \multicolumn{1}{c|}{HSE-P ↑} & \multicolumn{1}{c|}{HSE-T ↑} & \cellcolor[HTML]{D7D7D7}\textbf{HSE ↑}           \\ \hline
Baseline (N/A)                  & \multicolumn{1}{c|}{96.43} & \multicolumn{1}{c|}{93.04} & \cellcolor[HTML]{D7D7D7}\textbf{89.72}         & CLIP (ViT-H-14)             & \multicolumn{1}{c|}{96.86} & \multicolumn{1}{c|}{93.46} & \cellcolor[HTML]{D7D7D7}\textbf{90.52 \gain{+0.80}} \\ \hline
MAE (ViT-B)                     & \multicolumn{1}{c|}{96.89} & \multicolumn{1}{c|}{93.72} & \cellcolor[HTML]{D7D7D7}\textbf{90.81 \gain{+1.09}} & DINOv2 (ViT-S/14)               & \multicolumn{1}{c|}{96.66} & \multicolumn{1}{c|}{93.63} & \cellcolor[HTML]{D7D7D7}\textbf{90.51 \gain{+0.79}} \\
MAE (ViT-L)                     & \multicolumn{1}{c|}{96.89} & \multicolumn{1}{c|}{94.44} & \cellcolor[HTML]{D7D7D7}\textbf{91.51 \gain{+1.79}} & DINOv2 (ViT-B/14)               & \multicolumn{1}{c|}{96.91} & \multicolumn{1}{c|}{93.87} & \cellcolor[HTML]{D7D7D7}\textbf{90.98 \gain{+1.26}} \\ \cline{1-4}
SAM1 (ViT-L)                    & \multicolumn{1}{c|}{96.70} & \multicolumn{1}{c|}{93.36} & \cellcolor[HTML]{D7D7D7}\textbf{90.29 \gain{+0.57}} & DINOv2 (ViT-L/14)               & \multicolumn{1}{c|}{96.92} & \multicolumn{1}{c|}{94.21} & \cellcolor[HTML]{D7D7D7}\textbf{91.31 \gain{+1.59}} \\ \cline{5-8} 
SAM1 (ViT-H)                    & \multicolumn{1}{c|}{97.15} & \multicolumn{1}{c|}{93.39} & \cellcolor[HTML]{D7D7D7}\textbf{90.73 \gain{+1.01}} & DINOv3 (ViT-S+/16)              & \multicolumn{1}{c|}{96.68} & \multicolumn{1}{c|}{93.88} & \cellcolor[HTML]{D7D7D7}\textbf{90.76 \gain{+1.04}} \\ \cline{1-4}
SAM2.1 (Base+)                  & \multicolumn{1}{c|}{96.85} & \multicolumn{1}{c|}{93.98} & \cellcolor[HTML]{D7D7D7}\textbf{91.02 \gain{+1.30}} & DINOv3 (ViT-B/16)               & \multicolumn{1}{c|}{97.10} & \multicolumn{1}{c|}{94.00} & \cellcolor[HTML]{D7D7D7}\textbf{91.27 \gain{+1.55}} \\
SAM2.1 (Large)                  & \multicolumn{1}{c|}{96.24} & \multicolumn{1}{c|}{95.27} & \cellcolor[HTML]{D7D7D7}\textbf{91.69 \gain{+1.97}} & DINOv3 (ViT-L/16)               & \multicolumn{1}{c|}{96.63} & \multicolumn{1}{c|}{94.85} & \cellcolor[HTML]{D7D7D7}\textbf{91.66 \gain{+1.94}} \\ \hline
\end{tabular}
}
\end{minipage}
\vspace{-5pt}
\end{table}

\subsection{Ablation Experiment}
To fully validate the proposed FDEP framework, we perform detailed break-down ablation experiments. From Tab.~\ref{tab:tab03}, compared with the \textit{“Original”}, introducing the frozen VFM features improves the $HSE$ metric by 1.68, indicating that the global semantic priors provided by VFMs can effectively compensate for the sparsity and limited semantic representation of target features. At the same time, from \textit{“Scheme1-Scheme2”} and \textit{“Scheme3-Scheme4”}, the proposed SAMF module further enhances the performance gains brought by VFM feature integration, yielding an additional average 0.36 $HSE$ gain with negligible inference cost. In addition, from \textit{“Scheme1-Scheme3”} and \textit{“Scheme2-Scheme4”}, incorporating the proposed CO-ISD strategy significantly reduces inference time by 63.0\% (from 0.100 to 0.037) while maintaining only a negligible decrease in $HSE$, markedly enhancing inference efficiency. Furthermore, by comparing the \textit{“Original”} and \textit{“Scheme4”}, SIRST networks equipped with the FDEP framework achieve an $HSE$ improvement of 1.59 under the same inference time. In summary, the frozen VFM features provide strong global semantic priors, the SAMF module enables efficient multi-level semantic fusion, and the CO-ISD strategy ensures an optimal balance between performance and deployability. Together, these three components constitute the core driving forces behind the performance improvements achieved by the proposed FDEP framework. More detailed ablation experiments (VFM feature-level integration, VFM scale, SAMF module placement, etc.) and analysis can be found in the Appendix~\ref{sec:more ablation experiments}.

\subsection{Discussion}
\vspace{-6pt}
\label{sec:roc_limitations}
\textbf{\textit{1) The limitations of directly applying VFMs to SIRST detection.}} As shown in Tab.~\ref{tab:tab07}, naively transferring VFMs to the SIRST task performs poorly under the tested decoders. Specifically, directly connecting a VFM to a decoder or replacing the MSDA-Net encoder remains far below the FDEP-based method. For instance, the \textit{“VFMs + Decoder”} achieves an $HSE$ of only 21.77, and unfreezing the VFM improves it marginally to 23.03. This shows that although VFMs exhibit powerful representational capability for natural images, their global semantic features are not directly compatible with SIRST feature distribution, leading to a notable domain gap. Moreover, compared with the \textit{“DINO + Decoder”}, using \textit{“DINO + MSDA-Net (Decoder)”} yields a substantial improvement, indicating that task-specific network structures are better suited to capturing the subtle characteristics of infrared small targets. However, even under this configuration, it still fails to fully leverage the potential of VFMs.  In contrast, the proposed FDEP achieves adaptive transfer from foundation-level semantics to task-specific semantics through explicit semantic fusion and implicit knowledge transfer.
% The features produced by VFMs are inherently designed for natural scenes, and their feature space deviates considerably from the grayscale distribution, saliency patterns, and scale characteristics of infrared small targets.

\begin{table}[t]
\caption {Cross-dataset generalization evaluated by HSE. The first and second header rows indicate the training and testing datasets, respectively. All variants use the MSDA-Net as the detector.}
\centering
\label{tab:tabs-cross-datasets}
\setlength{\tabcolsep}{1.5mm}
\renewcommand{\arraystretch}{1.2}
\resizebox{1.0\linewidth}{!}{
\begin{tabular}{c|cc|cc|cc}
\hline
\multirow{2}{*}{Schemes} & \multicolumn{2}{c|}{NUAA-SIRST}                      & \multicolumn{2}{c|}{NUDT-SIRST}                    & \multicolumn{2}{c}{IRSTD-1K}                        \\ \cline{2-7} 
                         & \multicolumn{1}{c|}{NUDT-SIRST}     & IRSTD-1K       & \multicolumn{1}{c|}{NUAA-SIRST}    & IRSTD-1K      & \multicolumn{1}{c|}{NUAA-SIRST}    & NUDT-SIRST     \\ \hline
Original                 & \multicolumn{1}{c|}{48.70}          & 45.32          & \multicolumn{1}{c|}{63.12}         & 31.10         & \multicolumn{1}{c|}{75.39}         & 38.97          \\ \hline
+FDEP w/o CO-ISD         & \multicolumn{1}{c|}{61.35 \color[HTML]{7030A0}(+12.65)} & 62.61 \color[HTML]{7030A0}(+17.29) & \multicolumn{1}{c|}{68.21 \color[HTML]{7030A0}(+5.09)} & 38.17 \color[HTML]{7030A0}(+7.07) & \multicolumn{1}{c|}{80.46 \color[HTML]{7030A0}(+5.07)} & 53.32 \color[HTML]{7030A0}(+14.35) \\ \hline
+FDEP Framework          & \multicolumn{1}{c|}{52.26 \color[HTML]{7030A0}(+3.56)}  & 49.50 \color[HTML]{7030A0}(+4.18)  & \multicolumn{1}{c|}{67.23 \color[HTML]{7030A0}(+4.11)} & 31.41 \color[HTML]{7030A0}(+0.31) & \multicolumn{1}{c|}{83.34 \color[HTML]{7030A0}(+7.95)} & 41.98 \color[HTML]{7030A0}(+3.01)  \\ \hline
\end{tabular}
}
\vspace{-4pt}
\end{table}

% \begin{table}[t]
% \caption {Cross-dataset generalization evaluated by HSE. The first and second header rows indicate the training and testing datasets, respectively. All FDEP variants use MSDA-Net as the detector.}
% \centering
% \label{tab:tabs-cross-datasets}
% \setlength{\tabcolsep}{1.5mm}
% \renewcommand{\arraystretch}{1.2}
% \resizebox{1.0\linewidth}{!}{
% \begin{tabular}{c|cc|cc|cc}
% \hline
% \multirow{2}{*}{Schemes} & \multicolumn{2}{c|}{NUAA-SIRST}                      & \multicolumn{2}{c|}{NUDT-SIRST}                    & \multicolumn{2}{c}{IRSTD-1K}                        \\ \cline{2-7} 
%                          & \multicolumn{1}{c|}{NUDT-SIRST}     & IRSTD-1K       & \multicolumn{1}{c|}{NUAA-SIRST}    & IRSTD-1K      & \multicolumn{1}{c|}{NUAA-SIRST}    & NUDT-SIRST     \\ \hline
% Original                 & \multicolumn{1}{c|}{48.70}          & 45.32          & \multicolumn{1}{c|}{63.12}         & 31.10         & \multicolumn{1}{c|}{75.39}         & 38.97          \\ \hline
% +FDEP w/o CO-ISD         & \multicolumn{1}{c|}{61.35 \gain{+12.65}} & 62.61 \gain{+17.29} & \multicolumn{1}{c|}{68.21 \gain{+5.09}} & 38.17 \gain{+7.07} & \multicolumn{1}{c|}{80.46 \gain{+5.07}} & 53.32 \gain{+14.35} \\ \hline
% +FDEP Framework          & \multicolumn{1}{c|}{52.26 \gain{+3.56}}  & 49.50 \gain{+4.18}  & \multicolumn{1}{c|}{67.23 \gain{+4.11}} & 31.41 \gain{+0.31} & \multicolumn{1}{c|}{83.34 \gain{+7.95}} & 41.98 \gain{+3.01}  \\ \hline
% \end{tabular}
% }
% \vspace{-6pt}
% \end{table}

\begin{figure}[t]
  \captionsetup{skip=4pt}
  \centering
   \includegraphics[width=1.0\columnwidth]{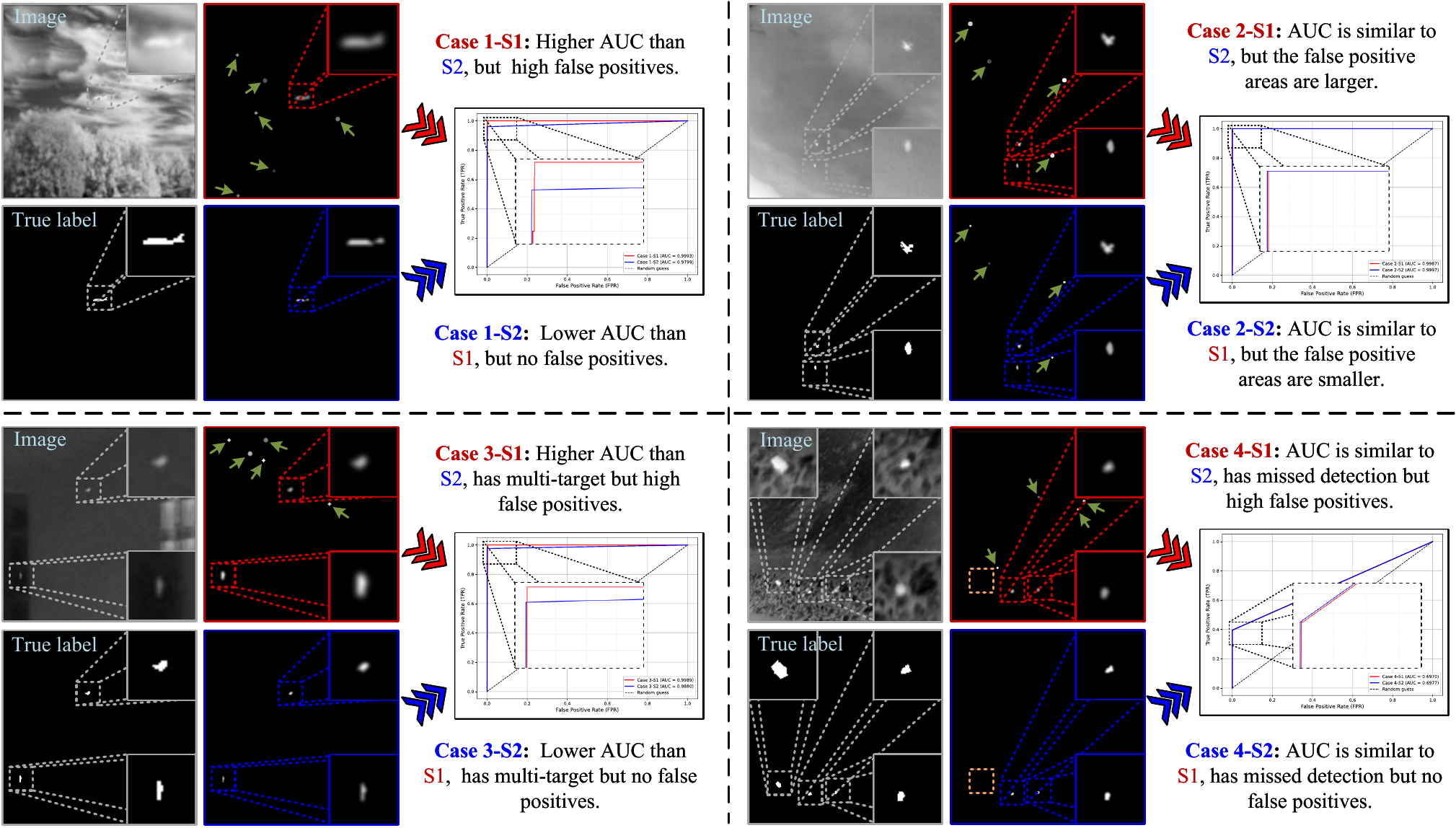}
   \caption{ROC comparisons across multiple different and representative cases. The \textcolor{red}{red} and \textcolor{blue}{blue} correspond to two prediction samples (S1 and S2), respectively. The \textcolor{orange}{orange boxes} indicate missed detections, and the \textcolor{palegreen}{green arrows} denote false detections.}
   \label{fig:figs-roc}
   \vspace{-6pt}
\end{figure}

\textbf{\textit{2) The limitations of using ROC curves for SIRST detection evaluation.}}
Conventional ROC curves fail to accurately reflect model performance in SIRST detection due to the extreme class imbalance. Specifically, target pixels occupy only a very small percentage of the image, while background pixels overwhelmingly dominate. Under such conditions, even if a model generates a large number of false positives, as long as the overall negative sample prediction is correct, the ROC curve will still present the illusion of a high AUC. The Cases in Fig.~\ref{fig:figs-roc} can effectively verify this. For example, Case I-S1 still attains a high AUC of 0.9993 despite many false positives, whereas Case I-S2 shows much better overall detection yet a lower AUC (0.9799). Tab.~\ref{tab:tabs-s05} in the Appendix~\ref{sec:more Discussion} provides more results.

\textbf{\textit{3) The FDEP can be extended to various VFMs.}} In addition to seamlessly adapting to existing encoder-decoder-based SIRST networks, our FDEP can also be easily adapted to various VFMs, demonstrating good versatility and scalability. From Tab.~\ref{tab:tab08}, introducing semantic priors from different VFMs into FDEP consistently yields stable gains. Specifically, compared to the \textit{``Baseline (N/A)''} without introducing VFMs, FDEP with different VFMs improves the \textit{HSE} metric by 0.57-1.97, verifying that the framework does not depend on a specific VFM, but can widely benefit from the high-level semantic representation of VFMs. In addition, Tab.~\ref{tab:tab08} also shows that higher \textit{HSE} often corresponds to larger VFM scales, indicating that richer semantic and structural representations can provide stronger transferable priors for FDEP.

\textbf{\textit{4) The role of VFM-derived semantic priors in SIRST detection.}} Although infrared small targets exhibit limited semantic appearance, the VFM-derived priors in FDEP provide complementary information about background structure, scene context, and mid-level visual patterns. Specifically, background structural priors capture regional continuity and structural regularities, providing a global reference for identifying local anomalies. Scene-level contextual priors encode spatial layouts and dependencies between regions, helping distinguish potential targets from bright background structures such as clouds and building edges. Mid-level visual representations capture generic edges, contours, and textures that complement local contrast and target boundary cues. Thus, the semantic priors primarily describe contextual and structural information rather than target category semantics. Through SAMF, these priors are aligned and fused with task-specific features to support small target discrimination and background suppression.

\textbf{\textit{5) Evaluation of Cross-Dataset Generalization.}} To evaluate generalization under dataset shifts, we conduct cross-dataset experiments on NUAA-SIRST, NUDT-SIRST, and IRSTD-1K, training on one dataset and testing on each of the other two. As shown in Tab.~\ref{tab:tabs-cross-datasets}, compared with the original MSDA-Net, the complete FDEP framework improves HSE by 0.31-7.95 across all six transfer settings without additional VFM-related inference overhead. The accuracy-oriented variant, ``+ FDEP w/o CO-ISD'', achieves gains of 5.07-17.29, generally providing stronger cross-dataset performance by retaining the VFM and SAMF during inference. These results suggest that frozen VFM representations provide transferable contextual and structural priors, while CO-ISD enables the lightweight branch to retain part of their benefits after removing the VFM and SAMF. Nevertheless, the gains vary across transfer settings, with a modest improvement of 0.31 for the complete FDEP when transferring from NUDT-SIRST to IRSTD-1K, indicating that substantial domain shifts remain challenging. The two variants therefore offer complementary choices for accuracy-oriented applications and efficient deployment.

\section{Conclusion}
\label{sec:conlusion}
This work rethinks the SIRST detection paradigm and proposes the FDEP framework, which can seamlessly adapt to diverse VFMs and SIRST networks. Our FDEP introduces frozen VFM representations as global semantic priors and employs a SAMF module to achieve feature alignment and semantic injection, thereby enhancing weak-target responses and suppressing complex-background interference. Meanwhile, a CO-ISD strategy is proposed, which achieves implicit transfer of semantic priors through parameter sharing and synchronous backpropagation. In addition, we construct the $HSE$ metric to address the redundancy and incompleteness of existing evaluation systems. Extensive experiments demonstrate that the SIRST networks equipped with the FDEP framework significantly improve detection accuracy while maintaining VFM-free inference efficiency. We hope this study draws attention to research on deep collaboration between VFMs and task-specific networks.

\section*{Acknowledgements}
\vspace{-2pt}
This work is partially supported by the National Natural Science Foundation of China (No. 62306261), HK RGC-Early Career Scheme (No. 24211525), ITSP Platform Project (No. ITS/600/24FP), the Centre for Perceptual and Interactive Intelligence, a CUHK-led InnoCentre under the InnoHK initiative of the Innovation and Technology Commission of the Hong Kong Special Administrative Region Government, Hong Kong RGC Strategic Topics Grant (No. STG1/E-403/24-N and No. STG1/M-404/26-N), CUHK-CUHK(SZ)-GDST Joint Collaboration Fund (No. YSP26-4760949), the LiaoNing Revitalization Program under Grant no. XLYC2201001, and the Institute Basic Research Program (No. E429040701).

% \newpage
\bibliographystyle{plainnat}
\bibliography{main}

%%%%%%%%%%%%%%%%%%%%%%%%%%%%%%%%%%%%%%%%%%%%%%%%%%%%%%%%%%%%
\newpage
\appendix

In this appendix, we provide additional details and results to complement the main paper. In Appendix~\ref{sec:more ablation experiments}, we present more ablation experiments to thoroughly explore the performance of the proposed FDEP framework. In Appendix~\ref{sec:compared wit more sota methods}, we offer quantitative comparisons with more existing methods. In Appendix~\ref{sec:more qualitative analyses}, we provide additional qualitative comparisons on three individual datasets. In Appendix~\ref{sec:more Discussion}, we provide further discussion. In Appendix~\ref{sec:broader_impacts}, we provide a discussion of the potential societal impacts and practical applications of our work. In Appendix~\ref{sec:limitations_and_future_work}, we provide the limitations of the current study and outline possible directions for future work.

\section{More Ablation Experiments}
\label{sec:more ablation experiments}
% In this section, we have added the investigation of VFMs feature-level integration, the investigation of VFMs scale and integration strategy, the investigation of SAMF module placement, and further investigation of CO-ISD strategy.

In this section, we present additional ablation studies to further investigate VFM feature-level integration, VFM scale and integration strategy, SAMF module placement, and the CO-ISD strategy.

\textbf{\textit{1) Investigation of VFM feature-level integration.}} We evaluate DINO feature-level integration on the SIRST3 dataset. From Tab.~\ref{tab:tab04}, compared with shallow-layer features (e.g., the 1$_{st}$ layer), deeper features yield more substantial and stable gains. Specifically, relative to the \textit{``Original''} in Tab.~\ref{tab:tab03}, introducing only the 1$_{st}$ layer improves the $HSE$ by 0.38 (from 89.72 to 90.10), while the 8$_{th}$, 16$_{th}$, and 24$_{th}$ layers achieve gains of 1.32 (from 89.72 to 91.04), 1.35 (from 89.72 to 91.07), and 1.26 (from 89.72 to 90.98), respectively. This indicates that high-level semantic features play a crucial role in enhancing robustness and suppressing complex backgrounds. In addition, multi-level feature fusion yields the best $HSE$ at layers 6/12/18/24, indicating complementary semantics across layers for hierarchical representations. Notably, adding more layers (“4/9/14/19/24”) may lead to feature-space redundancy and overlap, providing limited additional benefits.

\begin{table}[t]
\centering
\caption{Investigation of introducing different VFMs levels on the SIRST3 dataset.}
\setlength{\tabcolsep}{3.5mm}
\renewcommand{\arraystretch}{1.1}
\resizebox{\linewidth}{!}{
\begin{tabular}{c|ccc|c|ccc}
\hline
 & \multicolumn{3}{c|}{MSDA-Net + FDEP} & & \multicolumn{3}{c}{MSDA-Net + FDEP} \\ \cline{2-4} \cline{6-8}
\multirow{-2}{*}{Levels} &
\multicolumn{1}{c|}{HSE-P ↑} & \multicolumn{1}{c|}{HSE-T ↑} & HSE ↑ &
\multirow{-2}{*}{Levels} &
\multicolumn{1}{c|}{HSE-P ↑} & \multicolumn{1}{c|}{HSE-T ↑} & HSE ↑ \\ \hline
1  & \multicolumn{1}{c|}{96.17} & \multicolumn{1}{c|}{93.69} & 90.10 &
12/24 & \multicolumn{1}{c|}{96.70} & \multicolumn{1}{c|}{94.31} & 91.21 \\ \hline
8  & \multicolumn{1}{c|}{96.91} & \multicolumn{1}{c|}{93.95} & 91.04 &
8/16/24 & \multicolumn{1}{c|}{96.91} & \multicolumn{1}{c|}{94.19} & 91.28 \\ \hline
16 & \multicolumn{1}{c|}{96.68} & \multicolumn{1}{c|}{94.20} & 91.07 &
\cellcolor[HTML]{D7D7D7}6/12/18/24 &
\multicolumn{1}{c|}{\cellcolor[HTML]{D7D7D7}{\textbf{96.92}}} &
\multicolumn{1}{c|}{\cellcolor[HTML]{D7D7D7}94.21} &
\cellcolor[HTML]{D7D7D7}{\textbf{91.31}} \\ \hline
24 & \multicolumn{1}{c|}{96.92} & \multicolumn{1}{c|}{93.88} & 90.98 &
4/9/14/19/24 & \multicolumn{1}{c|}{96.41} & \multicolumn{1}{c|}{{\textbf{94.59}}} & 91.19 \\ \hline
\end{tabular}}
\label{tab:tab04}
\end{table}
\vspace{2pt}

\begin{table}[t]
\centering
\caption{Investigation of VFM scale and SAMF strategy in FDEP on the SIRST3 dataset.}
\setlength{\tabcolsep}{2mm}
\renewcommand{\arraystretch}{1.1}
\resizebox{\linewidth}{!}{
\begin{tabular}{c|c|ccc|c|ccc}
\hline
 & & \multicolumn{3}{c|}{MSDA-Net + FDEP} & & \multicolumn{3}{c}{MSDA-Net + FDEP} \\ \cline{3-5} \cline{7-9}
\multirow{-2}{*}{Scales} & \multirow{-2}{*}{SAMF} &
\multicolumn{1}{c|}{HSE-P ↑} & \multicolumn{1}{c|}{HSE-T ↑} & HSE ↑ &
\multirow{-2}{*}{SAMF} &
\multicolumn{1}{c|}{HSE-P ↑} & \multicolumn{1}{c|}{HSE-T ↑} & HSE ↑ \\ \hline
ViT-S/14 & \textbf{\ding{55}} & \multicolumn{1}{c|}{96.52} & \multicolumn{1}{c|}{93.45} & 90.20 &
\cellcolor[HTML]{D7D7D7}\textbf{\ding{51}} &
\multicolumn{1}{c|}{\cellcolor[HTML]{D7D7D7}{\textbf{96.66 \gain{+0.14}}}} &
\multicolumn{1}{c|}{\cellcolor[HTML]{D7D7D7}{\textbf{93.63 \gain{+0.18}}}} &
\cellcolor[HTML]{D7D7D7}{\textbf{90.51 \gain{+0.31}}} \\ \hline
ViT-B/14 & \textbf{\ding{55}} & \multicolumn{1}{c|}{96.49} & \multicolumn{1}{c|}{93.70} & 90.41 &
\cellcolor[HTML]{D7D7D7}\textbf{\ding{51}} &
\multicolumn{1}{c|}{\cellcolor[HTML]{D7D7D7}{\textbf{96.91 \gain{+0.42}}}} &
\multicolumn{1}{c|}{\cellcolor[HTML]{D7D7D7}{\textbf{93.87 \gain{+0.17}}}} &
\cellcolor[HTML]{D7D7D7}{\textbf{90.98 \gain{+0.57}}} \\ \hline
ViT-L/14 & \textbf{\ding{55}} & \multicolumn{1}{c|}{96.76} & \multicolumn{1}{c|}{93.73} & 90.92 &
\cellcolor[HTML]{D7D7D7}\textbf{\ding{51}} &
\multicolumn{1}{c|}{\cellcolor[HTML]{D7D7D7}{\textbf{96.92 \gain{+0.16}}}} &
\multicolumn{1}{c|}{\cellcolor[HTML]{D7D7D7}{\textbf{94.21 \gain{+0.48}}}} &
\cellcolor[HTML]{D7D7D7}{\textbf{91.31 \gain{+0.39}}} \\ \hline
\end{tabular}}
\label{tab:tab05}
\end{table}

\begin{table}[t]
\caption{Investigation of VFM scale and SAMF strategy in FDEP w/o CO-ISD on the SIRST3 dataset.}
\label{tab:tabs-s02}
\setlength{\tabcolsep}{2mm}
\renewcommand{\arraystretch}{1.1}
\resizebox{\columnwidth}{!}{
\begin{tabular}{c|c|ccc|c|ccc}
\hline
                         &                        & \multicolumn{3}{c|}{MSDA-Net + FDEP w/o CO-ISD}                     &                                    & \multicolumn{3}{c}{MSDA-Net + FDEP w/o CO-ISD}                                                                                                                                                                                          \\ \cline{3-5} \cline{7-9} 
\multirow{-2}{*}{Scales} & \multirow{-2}{*}{SAMF} & \multicolumn{1}{c|}{HSE-P ↑} & \multicolumn{1}{c|}{HSE-T ↑} & HSE ↑ & \multirow{-2}{*}{SAMF}             & \multicolumn{1}{c|}{HSE-P ↑}                                                       & \multicolumn{1}{c|}{HSE-T ↑}                                                       & HSE ↑                                                          \\ \hline
ViT-S/14                 & \textbf{\ding{55}}             & \multicolumn{1}{c|}{97.08}   & \multicolumn{1}{c|}{93.30}   & 90.58 & \cellcolor[HTML]{CFCECE}\textbf{\ding{51}} & \multicolumn{1}{c|}{\cellcolor[HTML]{CFCECE}{\color[HTML]{000000} \textbf{97.21 \gain{+0.13}}}} & \multicolumn{1}{c|}{\cellcolor[HTML]{CFCECE}{\color[HTML]{000000} \textbf{93.61 \gain{+0.31}}}} & \cellcolor[HTML]{CFCECE}{\color[HTML]{000000} \textbf{91.00 \gain{+0.42}}} \\ \hline
ViT-B/14                 & \textbf{\ding{55}}             & \multicolumn{1}{c|}{96.76}   & \multicolumn{1}{c|}{94.08}   & 91.03 & \cellcolor[HTML]{CFCECE}\textbf{\ding{51}} & \multicolumn{1}{c|}{\cellcolor[HTML]{CFCECE}{\color[HTML]{000000} \textbf{97.34 \gain{+0.58}}}} & \multicolumn{1}{c|}{\cellcolor[HTML]{CFCECE}{\color[HTML]{000000} \textbf{93.83 \gain{-0.25}}}} & \cellcolor[HTML]{CFCECE}{\color[HTML]{000000} \textbf{91.34 \gain{+0.31}}} \\ \hline
ViT-L/14                 & \textbf{\ding{55}}             & \multicolumn{1}{c|}{96.93}   & \multicolumn{1}{c|}{94.29}   & 91.40 & \cellcolor[HTML]{CFCECE}\textbf{\ding{51}} & \multicolumn{1}{c|}{\cellcolor[HTML]{CFCECE}{\color[HTML]{000000} \textbf{97.25 \gain{+0.32}}}} & \multicolumn{1}{c|}{\cellcolor[HTML]{CFCECE}{\color[HTML]{000000} \textbf{94.32 \gain{+0.03}}}} & \cellcolor[HTML]{CFCECE}{\color[HTML]{000000} \textbf{91.73 \gain{+0.33}}} \\ \hline
\end{tabular}
}
% \vspace{-3pt}
\end{table}

\begin{table}[!t]
\caption {Investigation of SAMF placement on the SIRST3 dataset. \textit{``Baseline''} denotes the original MSDA-Net. \textit{``Each stage''} indicates injecting VFM features via SAMF at each stage. \textit{CL}: Cross-layer.}
\label{tab:tabs-s01}
\setlength{\tabcolsep}{2.0mm}
\renewcommand{\arraystretch}{1.05}
\resizebox{\linewidth}{!}{
\begin{tabular}{c|cccc|ccc}
\hline
                                    & \multicolumn{4}{c|}{Conventional Metrics}                                                                                                                                                       & \multicolumn{3}{c}{HSE Series Metrics}                                                                                                             \\ \cline{2-8} 
\multirow{-2}{*}{Scheme}            & \multicolumn{1}{c|}{mIoU ↑}                                  & \multicolumn{1}{c|}{nIoU ↑}                                  & \multicolumn{1}{c|}{P$_d$ ↑}                           & F$_a$ ↓            & \multicolumn{1}{c|}{HSE-P ↑}                                 & \multicolumn{1}{c|}{HSE-T ↑}                                 & HSE ↑                   \\ \hline
Baseline (MSDA-Net)                            & \multicolumn{1}{c|}{82.86}                                  & \multicolumn{1}{c|}{85.46}                                  & \multicolumn{1}{c|}{97.01}                         & 12.53          & \multicolumn{1}{c|}{96.43}                                  & \multicolumn{1}{c|}{93.04}                                  & 89.72                  \\ \hline
Encoder (Each stage)                & \multicolumn{1}{c|}{82.71}                                  & \multicolumn{1}{c|}{85.13}                                  & \multicolumn{1}{c|}{\textbf{97.81}}                & 15.20          & \multicolumn{1}{c|}{96.31 \gain{-0.12}}                                  & \multicolumn{1}{c|}{93.47 \gain{+0.43}}                                  & 90.02 \gain{+0.30}          \\
Decoder (Each stage)                & \multicolumn{1}{c|}{83.86}                                  & \multicolumn{1}{c|}{85.55}                                  & \multicolumn{1}{c|}{96.54}                         & \textbf{11.49} & \multicolumn{1}{c|}{96.80 \gain{+0.37}}                                  & \multicolumn{1}{c|}{93.27 \gain{+0.23}}                                  & 90.28 \gain{+0.56}          \\
CL Connection (Each stage) & \multicolumn{1}{c|}{83.23}                                  & \multicolumn{1}{c|}{85.81}                                  & \multicolumn{1}{c|}{97.28}                         & 21.71          & \multicolumn{1}{c|}{96.58 \gain{+0.15}}                                  & \multicolumn{1}{c|}{93.52 \gain{+0.48}}                                  & 90.33 \gain{+0.61}          \\ \hline
\rowcolor[HTML]{CFCECE} 
\textbf{Final FDEP (Ours)}                & \multicolumn{1}{c|}{\cellcolor[HTML]{CFCECE}\textbf{84.05}} & \multicolumn{1}{c|}{\cellcolor[HTML]{CFCECE}\textbf{86.33}} & \multicolumn{1}{c|}{\cellcolor[HTML]{CFCECE}97.54} & 14.24          & \multicolumn{1}{c|}{\cellcolor[HTML]{CFCECE}\textbf{96.92 \gain{+0.49}}} & \multicolumn{1}{c|}{\cellcolor[HTML]{CFCECE}\textbf{94.21 \gain{+1.17}}} & \textbf{91.31 \gain{+1.59}} \\ \hline
\end{tabular}
}
% \vspace{-10pt}
\end{table}

\begin{table}[!t]
\centering
\caption{Investigation of CO-ISD strategy on the SIRST3 dataset. \textit{EDL}: Explicit Distillation Loss.}
\setlength{\tabcolsep}{1.5mm}
\renewcommand{\arraystretch}{1.1}
\resizebox{\linewidth}{!}{
\begin{tabular}{c|c|ccc|c|ccc}
\hline
 & & \multicolumn{3}{c|}{MSDA-Net + FDEP} & & \multicolumn{3}{c}{MSDA-Net + FDEP} \\ \cline{3-5} \cline{7-9}
\multirow{-2}{*}{Share} & \multirow{-2}{*}{EDL} &
\multicolumn{1}{c|}{HSE-P ↑} & \multicolumn{1}{c|}{HSE-T ↑} & HSE ↑ &
\multirow{-2}{*}{EDL} &
\multicolumn{1}{c|}{HSE-P ↑} & \multicolumn{1}{c|}{HSE-T ↑} & HSE ↑ \\ \hline
Encoder (E)   & \textbf{\ding{51}} & \multicolumn{1}{c|}{94.71 \gain{+1.88}} & \multicolumn{1}{c|}{90.36 \gain{+3.30}} & 85.58 \gain{+4.89} &
\textbf{\ding{55}} & \multicolumn{1}{c|}{96.00 \gain{+0.92}} & \multicolumn{1}{c|}{92.50 \gain{+1.71}} & 88.81 \gain{+2.50} \\ \hline
Decoder (D)    & \textbf{\ding{51}} & \multicolumn{1}{c|}{94.67 \gain{+1.92}} & \multicolumn{1}{c|}{89.13 \gain{+4.53}} & 84.38 \gain{+6.09} &
\textbf{\ding{55}} & \multicolumn{1}{c|}{96.40 \gain{+0.52}} & \multicolumn{1}{c|}{91.96 \gain{+2.25}} & 88.64 \gain{+2.67}\\ \hline
\rowcolor[HTML]{D7D7D7}
E + D & \textbf{\ding{51}} &
\multicolumn{1}{c|}{\cellcolor[HTML]{D7D7D7}96.59} &
\multicolumn{1}{c|}{\cellcolor[HTML]{D7D7D7}93.66} &
90.47 &
\textbf{\ding{55}} &
\multicolumn{1}{c|}{\cellcolor[HTML]{D7D7D7}{\textbf{96.92}}} &
\multicolumn{1}{c|}{\cellcolor[HTML]{D7D7D7}{\textbf{94.21}}} &
{\textbf{91.31}} \\ \hline
\end{tabular}}
\label{tab:tab06}
\end{table}

\textbf{\textit{2) Investigation of VFM scale and integration strategy.}} From Tab.~\ref{tab:tab05}, as the scale of the VFMs increases, the model exhibits a consistent improvement in the $HSE$. This indicates that larger-scale VFMs provide richer global priors for SIRST detection. Meanwhile, compared to not using the SAMF module, using this module can achieve further performance improvements across all settings. Specifically, the $HSE$ metric improves by 0.31 (from 90.20 to 90.51) - 0.57 (from 90.41 to 90.98), demonstrating that the SAMF module effectively aligns the semantics from the VFMs with task-specific features, thereby achieving more efficient semantic fusion. To further investigate the impact of different scales of VFMs and their feature integration methods on the performance, in addition to the exploration on the \textit{``+ FDEP Framework''}, we also conduct detailed experimental studies on the high-precision \textit{``+ FDEP w/o CO-ISD''} framework. The experimental results are shown in Tab.~\ref{tab:tabs-s02}. Consistent with the experiments conducted using \textit{``+ FDEP Framework''}, the \textit{HSE} continues to improve as the scale of the VFMs increases. This indicates that larger backbone models possess stronger semantic abstraction and feature representation capabilities, thereby providing richer global priors for SIRST detection. Specifically, compared with \textit{``ViT-S/14''}, using \textit{``ViT-B/14''} improves \textit{HSE} by 0.45 (from 90.58 to 91.03) and 0.34 (from 91.00 to 91.34). Using \textit{``ViT-L/14''} improves \textit{HSE} by 0.82 (from 90.58 to 91.40) and 0.73 (from 91.00 to 91.73). In addition, compared with the variants without the SAMF module, equipping the model with the proposed SAMF module consistently yields performance gains across all backbone scales. Specifically, using SAMF improves \textit{HSE} by 0.31 (from 91.03 to 91.34) - 0.42 (from 90.58 to 91.00). These results further validate that the SAMF module can effectively align the semantic features of the VFMs with task-specific representations, enabling more efficient semantic transfer and fusion.

\textbf{\textit{3) Investigation of SAMF module placement.}} To systematically evaluate the impact of SAMF injection location on performance, we deployed it in each stage in the encoder, decoder, and cross-layer connections of MSDA-Net. The experimental results are shown in Tab.~\ref{tab:tabs-s01}. Overall, compared to the baseline without VFM, all three locations (encoder, decoder, and cross-layer connections) provide stable gains, improving \textit{HSE} by 0.30 (from 89.72 to 90.02), 0.56 (from 89.72 to 90.28), and 0.61  (from 89.72 to 90.33) respectively. The proposed FDEP further improves \textit{HSE} by 1.59  (from 89.72 to 91.31), achieving the best performance. This shows that VFMs priors have universal validity, but the injection location determines the upper limit of gain. Specifically, although the encoder's stage-by-stage injection improves P$_d$ and slightly increases \textit{HSE}-\textit{T}, it is accompanied by a slight decrease in mIoU/nIoU and an increase in F$_a$. This indicates that introducing too strong semantic priors in the early layers, which are dominated by local textures, can easily activate false background responses, thereby weakening the overall gains. In contrast, the stage-by-stage injection of the decoder is more conducive to reconstruction and denoising (F$_a$: 12.53→11.49, mIoU: 82.86→83.86), thus bringing a more significant improvement in \textit{HSE} (+0.56). However, P$_d$ slightly decreases, indicating that its main contribution is in refinement rather than recall enhancement. Cross-layer connections and staged injection can improve target-level stability (\textit{HSE}-\textit{T} is significantly improved, \textit{HSE} +0.61), but F$_a$ surges to 21.71, reflecting that direct superposition of semantic modulation in skip paths easily amplifies high-frequency background interference. Ultimately, FDEP simultaneously improved mIoU/nIoU and P$_d$ while keeping F$_a$ at 14.24, achieving a maximum \textit{HSE} of 91.31 (+1.59), further validating that its advantage stems from a more reasonable semantic alignment and fusion paradigm, rather than simple position stacking.

\begin{table*}[!t]

\caption{Performance comparison of different methods on the SIRST3 dataset. \textit{``+ FDEP w/o CO-ISD''} denotes the FDEP framework without using the CO-ISD strategy. \textcolor{darkred}{Red} denotes the best result, and \textcolor{blue}{blue} denotes the second best result. The values in parentheses denote the difference between each method and the \textit{``MSDA-Net + FDEP''}. The order of magnitude of F$_a$ is $10^{\scalebox{0.6}{$-6$}}$.}
\label{tab:tabs-s03}
\setlength{\tabcolsep}{1mm}
\renewcommand{\arraystretch}{1.3}
\resizebox{\linewidth}{!}{
\begin{tabular}{c|c|cccc|ccc|cc}
\hline
                                  &                                           & \multicolumn{4}{c|}{Conventional Accuracy Metrics}                                                                                                                                                                                                                                                  & \multicolumn{3}{c|}{HSE Series Accuracy Metrics}                                                                                                                                                                                                                & \multicolumn{2}{c}{Computational Cost}                                                            \\ \cline{3-11} 
\multirow{-2}{*}{Methods}         & \multirow{-2}{*}{Description}             & \multicolumn{1}{c|}{IoU ↑}                                                         & \multicolumn{1}{c|}{nIoU ↑}                                                        & \multicolumn{1}{c|}{P$_d$ ↑}                                                          & F$_a$ ↓                                 & \multicolumn{1}{c|}{HSE-P ↑}                                                               & \multicolumn{1}{c|}{HSE-T ↑}                                                               & HSE ↑                                                                 & \multicolumn{1}{c|}{Param (M)}                            & Time (s)                              \\
\hline
\end{tabular}
}
\end{table*}

\begin{table*}[t]
\caption{Performance comparison of different methods on the NUAA-SIRST, NUDT-SIRST, and IRSTD-1k datasets. \textit{``+ FDEP w/o CO-ISD''} denotes the FDEP without CO-ISD strategy. \textcolor{darkred}{Red} denotes the best result, and \textcolor{blue}{blue} denotes the second best result.}
\label{tab:tabs-s04}
\setlength{\tabcolsep}{1mm}
\renewcommand{\arraystretch}{1.4}
\resizebox{\linewidth}{!}{
\begin{tabular}{c|c|ccccc|ccccc|ccccc}
\hline
                           &                                           & \multicolumn{5}{c|}{NUAA-SIRST (213:214)}                                                                                                                                                                                                                                                                                                                                                                         & \multicolumn{5}{c|}{NUDT-SIRST (663:664)}                                                                                                                                                                                                                                                                                                                                                                         & \multicolumn{5}{c}{IRSTD-1K (800: 201)}                                                                                                                                                                                                                                                                                                                                          \\
\hline
\end{tabular}
}
\end{table*}

% \textbf{\textit{4) Investigation of CO-ISD strategy.}} We evaluate different parameter-sharing schemes and the explicit distillation loss. From Tab.~\ref{tab:tab06}, when only the encoder is shared, the absence of gradient collaboration in decoding limits the $HSE$ to 88.81. When only the decoder is shared, the absence of global constraints at the feature level limits the $HSE$ to 88.64. In contrast, sharing both encoder and decoder leads to a notable performance improvement, indicating that end-to-end gradient collaboration within a unified parameter space is crucial for effective implicit semantic transfer. In addition, the results using distillation loss show that adding an explicit distillation loss provides no benefit and slightly degrades performance. This indicates that explicit distillation tends to destabilize optimization in the extremely imbalanced SIRST task, whereas our CO-ISD strategy enables stable and efficient knowledge transfer. Meanwhile, these results suggest that CO-ISD is not a conventional teacher-student distillation process, but a gradient-level collaborative optimization strategy where VFM-induced semantics are transferred through shared deployable parameters.

\textbf{\textit{4) Investigation of CO-ISD strategy.}} We evaluate different parameter-sharing schemes and the explicit distillation loss. From Tab.~\ref{tab:tab06}, when only the encoder is shared, the absence of gradient collaboration in decoding limits the $HSE$ to 88.81. When only the decoder is shared, the absence of global constraints at the feature level limits the $HSE$ to 88.64. In contrast, sharing both encoder and decoder leads to a notable performance improvement, indicating that end-to-end gradient collaboration within a unified parameter space is crucial for effective implicit semantic transfer. In addition, the results using distillation loss show that adding an explicit distillation loss provides no benefit and slightly degrades performance. This indicates that explicit distillation tends to destabilize optimization in the extremely imbalanced SIRST task, whereas our CO-ISD strategy enables stable and efficient knowledge transfer. Meanwhile, these results suggest that CO-ISD is not a conventional teacher-student distillation process, but a gradient-level collaborative optimization strategy where VFM-induced semantics are transferred through shared deployable parameters. To further examine the interaction between the two branches, we retrain MSDA-Net equipped with FDEP on NUAA-SIRST and analyze their gradients with respect to the shared encoder-decoder parameters. The cosine similarity between the branch gradients increases from 0.550 in early training to 0.842 in late training, while their gradient norms remain of the same order of magnitude. These results suggest that, under the shared ground-truth detection objective, both branches contribute to shared parameter updates and gradually develop more consistent optimization directions. Together with the parameter-sharing ablations, they provide complementary optimization-level evidence supporting the collaborative training mechanism of CO-ISD.

\begin{figure*}[]
  \captionsetup{skip=4pt}
  \centering
   \includegraphics[width=\linewidth]{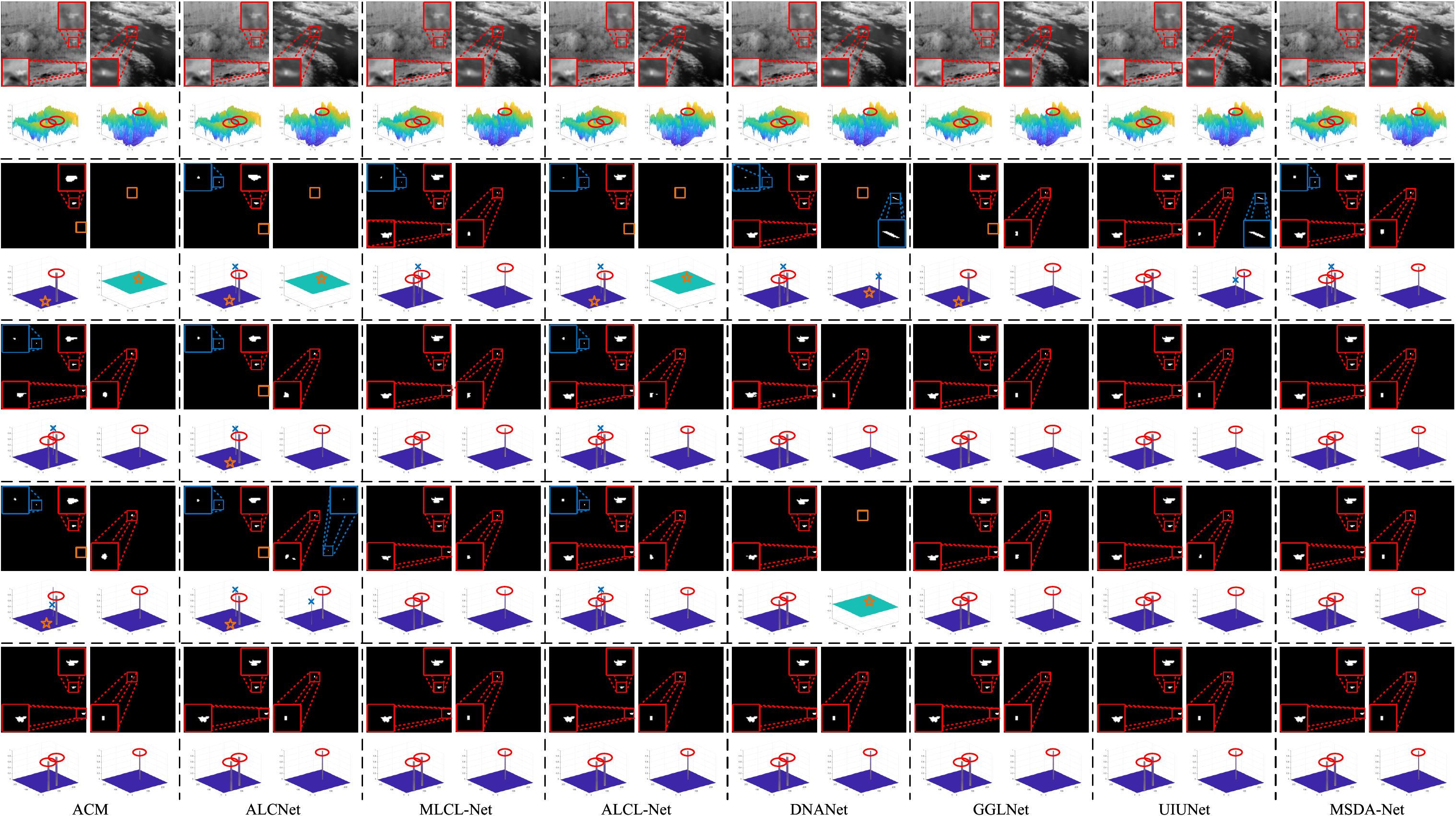}
   \caption{Visualization of several excellent methods on the SIRST3 dataset. \textcolor{red}{Red}: correct detections. \textcolor{Baselineblue}{Blue}: false positives. \textcolor{orange}{Orange}: missed detections. Every two rows from top to bottom: \textit{Image}, \textit{“Original”}, \textit{“+ FDEP w/o CO-ISD”}, \textit{“+ FDEP Framework”}, \textit{True label}.}
   \label{fig:figs-method-results}
   \vspace{-8pt}
\end{figure*}

\begin{figure*}[t]
  \captionsetup{skip=4pt}
  \centering
  \vspace{5pt}
   \includegraphics[width=\linewidth]{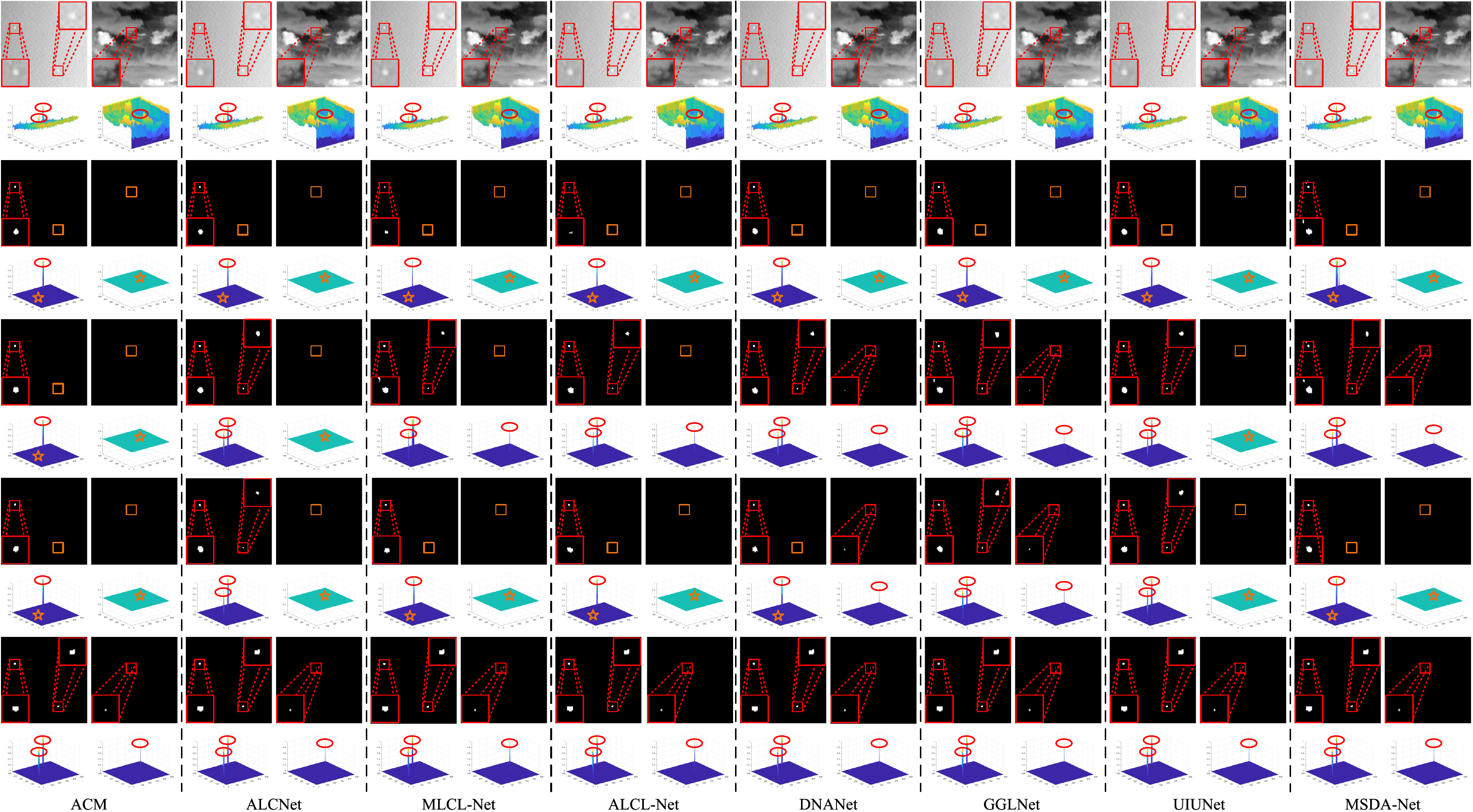}
   \caption{Visualization of some excellent methods on the NUAA-SIRST dataset. \textcolor{red}{Red}: correct detections; \textcolor{Baselineblue}{blue}: false positives; \textcolor{orange}{orange}: missed detections. Every two rows from top to bottom: \textit{Image}, \textit{``Original''}, \textit{``+ FDEP w/o CO-ISD''}, \textit{``+ FDEP Framework''}, \textit{True label}.}
   \label{fig:figs-s01}
\vspace{-10pt}
\end{figure*}

\section{Compared with more SOTA Methods}
\label{sec:compared wit more sota methods}
To fully validate the performance of MSDA-Net equipped with the FDEP framework, we conduct detailed quantitative comparisons against a series of recent excellent methods on the SIRST3~\cite{yu2025easy,ying2023mapping}, NUAA-SIRST~\cite{dai2021asymmetric}, NUDT-SIRST~\cite{li2022dense}, and IRSTD-1K~\cite{zhang2022isnet} datasets. In addition, we include three representative non-deep learning-based methods~\cite{qin2019infrared,wei2016multiscale,dai2017non} to ensure the completeness and rigor of the evaluation.

\begin{figure*}[t]
  \captionsetup{skip=4pt}
  \centering
   \includegraphics[width=\linewidth]{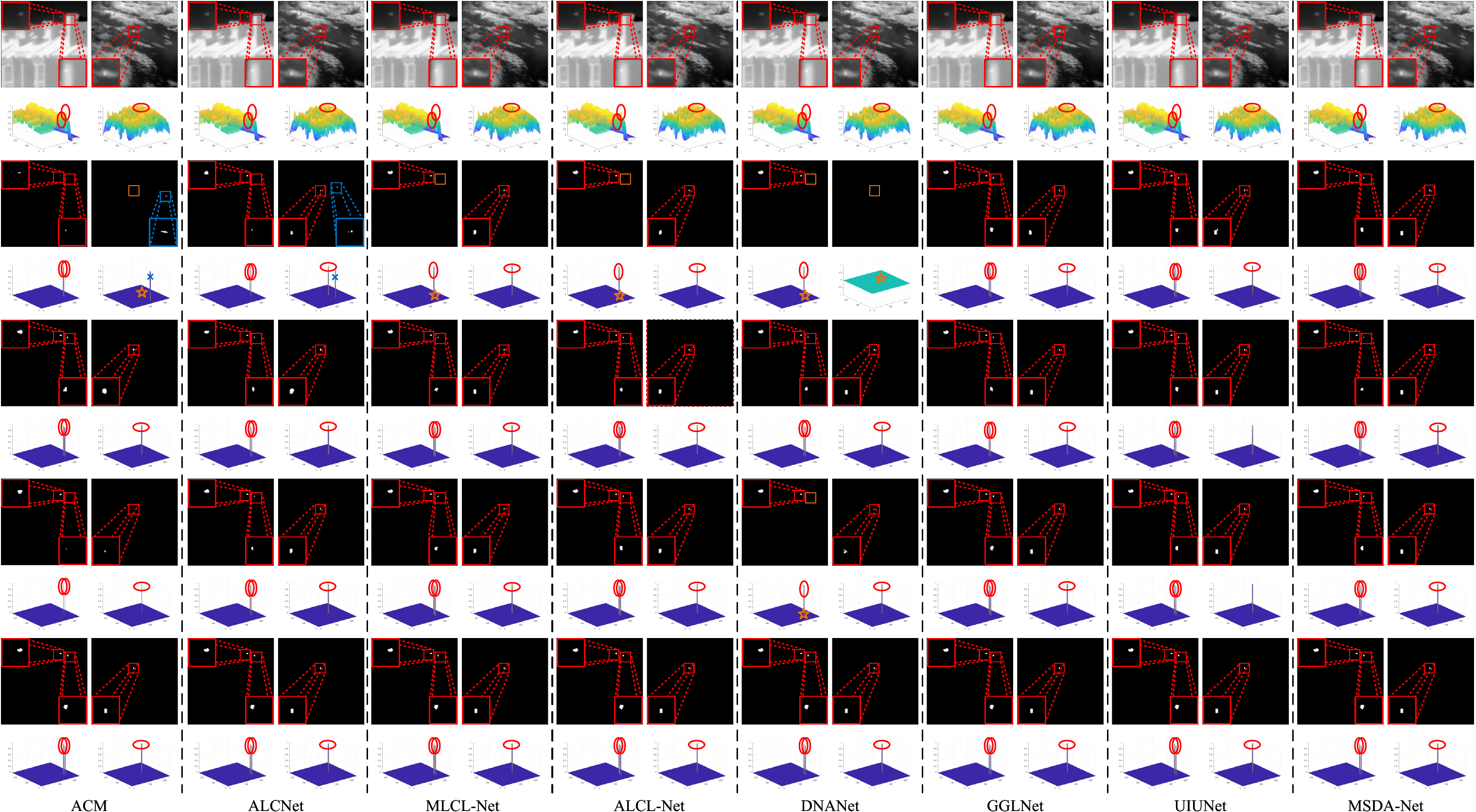}
   \caption{Visualization of some excellent methods on the NUDT-SIRST dataset. \textcolor{red}{Red}: correct detections; \textcolor{Baselineblue}{blue}: false positives; \textcolor{orange}{orange}: missed detections. Every two rows from top to bottom: \textit{Image}, \textit{``Original''}, \textit{``+ FDEP w/o CO-ISD''}, \textit{``+ FDEP Framework''}, \textit{True label}.}
   \label{fig:figs-s02}
    \vspace{-6pt}
\end{figure*}

\begin{figure*}[t]
  \captionsetup{skip=4pt}
  \centering
   \includegraphics[width=\linewidth]{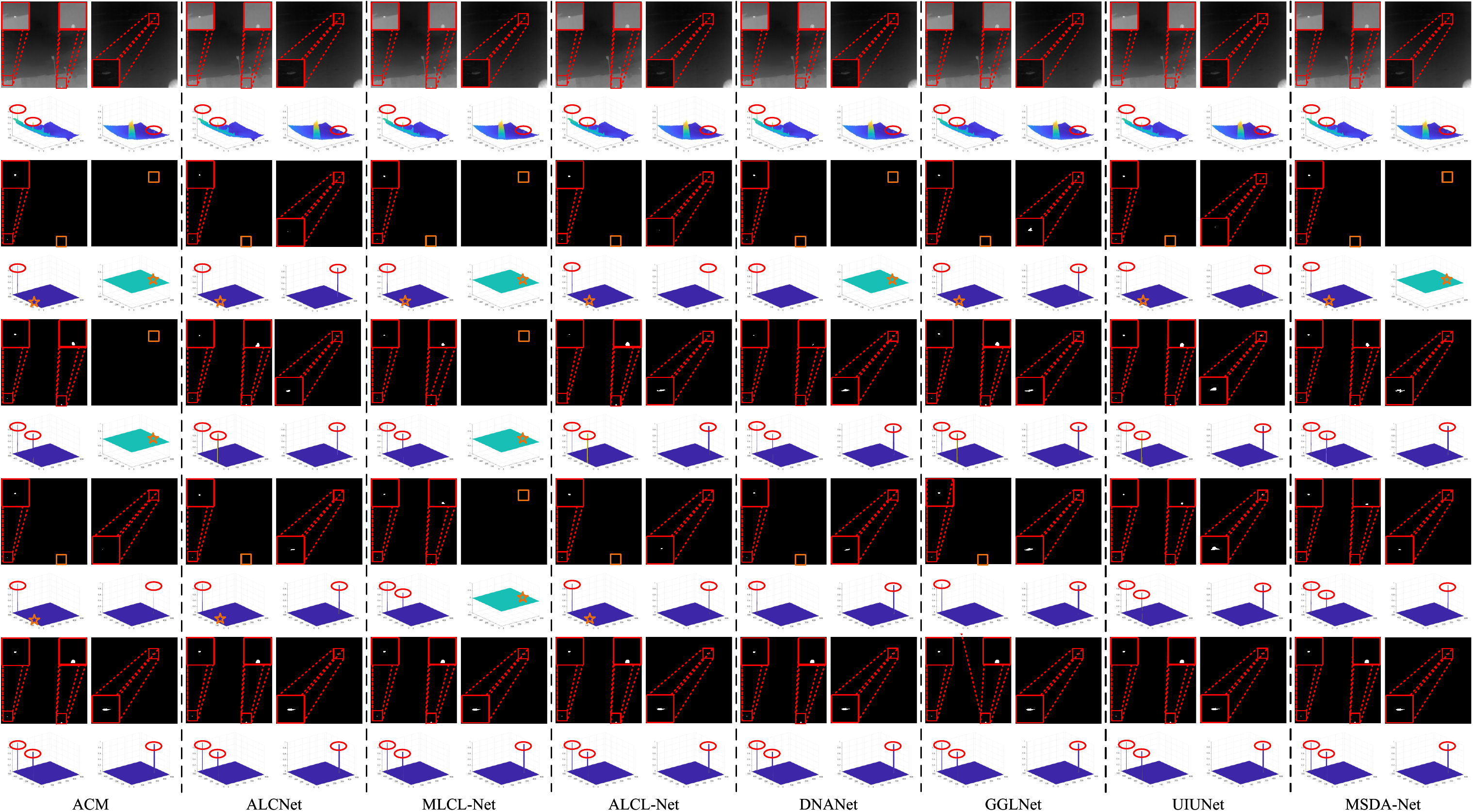}
   \caption{Visualization of some excellent methods on the IRSTD-1K dataset. \textcolor{red}{Red}: correct detections; \textcolor{Baselineblue}{blue}: false positives; \textcolor{orange}{orange}: missed detections. Every two rows from top to bottom: \textit{Image}, \textit{``Original''}, \textit{``+ FDEP w/o CO-ISD''}, \textit{``+ FDEP Framework''}, \textit{True label}.}
   \label{fig:figs-s03}
    \vspace{-8pt}
\end{figure*}

\textbf{\textit{1) Evaluation on the SIRST3 dataset.}} The experimental results are presented in Tab.~\ref{tab:tabs-s03}. First, compared with traditional non-deep learning-based methods, deep learning-based SIRST detection networks achieve significantly better performance. Second, compared to other non-MSDA-Net methods, MSDA-Net equipped with the FDEP framework achieves state-of-the-art (SOTA) performance. Specifically, it improves \textit{HSE-P} by 0.92 (from 96.00 to 96.92) - 76.44 (from 20.48 to 96.92), improves \textit{HSE-T} by 0.80 (from 93.41 to 94.21) - 80.71 (from 13.50 to 94.21), and improves the overall \textit{HSE} by 2.21 (from 89.10 to 91.31) - 87.07 (from 4.24 to 91.31). Third, compared with MSDA-Net using \textit{``+ FDEP Framework''}, the variant with \textit{``+ FDEP w/o CO-ISD''} achieves higher accuracy, making it more suitable for scenarios requiring high precision. Although \textit{``+ FDEP Framework''} yields slightly lower accuracy, it accelerates inference by 65\% (from 0.100 s to 0.035 s) compared with \textit{``+ FDEP w/o CO-ISD''}. Finally, compared with the original MSDA-Net, \textit{``+ FDEP Framework''} improves \textit{HSE} by 1.59 (from 89.72 to 91.31) without increasing inference cost, fully demonstrating the strong performance of the FDEP framework.

\textbf{\textit{2) Evaluation on three individual datasets.}} From Tab.~\ref{tab:tabs-s04}, compared to other non-MSDA-Net methods, MSDA-Net equipped with the FDEP framework achieves SOTA performance on both NUDT-SIRST and IRSTD-1K datasets. At the same time, although original MSDA-Net performs consistently well on NUDT-SIRST and IRSTD-1K, its performance on NUAA-SIRST is relatively weaker. However, after being equipped with \textit{``+ FDEP w/o CO-ISD''}, MSDA-Net obtains a much more notable performance gain on NUAA-SIRST. This is because the original MSDA-Net lacks sufficient robustness with extremely limited training samples, whereas incorporating the semantic features from VFMs can effectively mitigate this issue. These results further demonstrate that systematically leveraging the semantic priors of VFMs is beneficial for improving SIRST detection performance, especially in challenging or extreme scenarios, highlighting the importance of integrating VFMs into the SIRST task. In addition, compared with the original MSDA-Net, both \textit{``+ FDEP w/o CO-ISD''} and \textit{``+ FDEP Framework''} yield significant improvements in the overall \textit{HSE} metric, further validating the robustness of the proposed FDEP framework in few-shot scenarios.

\begin{table}[t]
\caption {Quantitative comparison of multiple cases in Fig.~\ref{fig:figs-roc} of the Main Text.}
\centering
\label{tab:tabs-s05}
\setlength{\tabcolsep}{4mm}
\renewcommand{\arraystretch}{1.1}
\resizebox{1.0\linewidth}{!}{
\begin{tabular}{c|cc|cc|cc|cc}
\hline
                        & \multicolumn{2}{c|}{Case 1}                                                                                                & \multicolumn{2}{c|}{Case 2}                                                                                               & \multicolumn{2}{c|}{Case 3}                                                                                                & \multicolumn{2}{c}{Case 4}                                                                                                \\ \cline{2-9} 
\multirow{-2}{*}{Cases} & \multicolumn{1}{c|}{{\color[HTML]{C00000} S1}}                                     & {\color[HTML]{0000F9} S2}             & \multicolumn{1}{c|}{{\color[HTML]{C00000} S1}}                                    & {\color[HTML]{0000F9} S2}             & \multicolumn{1}{c|}{{\color[HTML]{C00000} S1}}                                     & {\color[HTML]{0000F9} S2}             & \multicolumn{1}{c|}{{\color[HTML]{C00000} S1}}                                    & {\color[HTML]{0000F9} S2}             \\ \hline
AUC ↑                     & \multicolumn{1}{c|}{0.9993}                                                        & 0.9799                                & \multicolumn{1}{c|}{0.9987}                                                       & 0.9997                                & \multicolumn{1}{c|}{0.9989}                                                        & 0.9880                                & \multicolumn{1}{c|}{0.6970}                                                       & 0.6977                                \\ \hline
hIoU (\%) ↑                    & \multicolumn{1}{c|}{{\color[HTML]{C00000} 16.67}}                                  & {\color[HTML]{0000F9} 19.00}          & \multicolumn{1}{c|}{{\color[HTML]{C00000} 27.43}}                                 & {\color[HTML]{0000F9} 27.43}          & \multicolumn{1}{c|}{{\color[HTML]{C00000} 18.87}}                                  & {\color[HTML]{0000F9} 38.10}          & \multicolumn{1}{c|}{{\color[HTML]{C00000} 15.93}}                                 & {\color[HTML]{0000F9} 38.52}          \\ \hline
\rowcolor[HTML]{D7D7D7} 
HSE-P (\%) ↑                   & \multicolumn{1}{c|}{\cellcolor[HTML]{D7D7D7}45.37}                                 & 89.75                                 & \multicolumn{1}{c|}{\cellcolor[HTML]{D7D7D7}14.24}                                & 36.25                                 & \multicolumn{1}{c|}{\cellcolor[HTML]{D7D7D7}31.95}                                 & 91.27                                 & \multicolumn{1}{c|}{\cellcolor[HTML]{D7D7D7}17.04}                                & 64.46                                 \\ \hline
\rowcolor[HTML]{D7D7D7} 
HSE-T (\%) ↑                   & \multicolumn{1}{c|}{\cellcolor[HTML]{D7D7D7}33.33}                                 & 100.00                                & \multicolumn{1}{c|}{\cellcolor[HTML]{D7D7D7}36.67}                                & 36.67                                 & \multicolumn{1}{c|}{\cellcolor[HTML]{D7D7D7}45.00}                                 & 100.00                                & \multicolumn{1}{c|}{\cellcolor[HTML]{D7D7D7}21.67}                                & 66.67                                 \\ \hline
\rowcolor[HTML]{D7D7D7} 
\textbf{HSE (\%) ↑}            & \multicolumn{1}{c|}{\cellcolor[HTML]{D7D7D7}{\color[HTML]{C00000} \textbf{15.12}}} & {\color[HTML]{0000F9} \textbf{89.75}} & \multicolumn{1}{c|}{\cellcolor[HTML]{D7D7D7}{\color[HTML]{C00000} \textbf{5.22}}} & {\color[HTML]{0000F9} \textbf{13.29}} & \multicolumn{1}{c|}{\cellcolor[HTML]{D7D7D7}{\color[HTML]{C00000} \textbf{14.39}}} & {\color[HTML]{0000F9} \textbf{91.27}} & \multicolumn{1}{c|}{\cellcolor[HTML]{D7D7D7}{\color[HTML]{C00000} \textbf{3.70}}} & {\color[HTML]{0000F9} \textbf{42.97}} \\ \hline
\end{tabular}
}
\vspace{-6pt}
\end{table}

\section{More Qualitative Analyses}
\label{sec:more qualitative analyses}
To more intuitively demonstrate the effectiveness of the proposed FDEP framework under mixed or few-shot scenarios, we visualize detection results on a mixed dataset (SIRST3) and three individual datasets (NUAA-SIRST, NUDT-SIRST, and IRSTD-1K). These data cover diverse imaging conditions and target characteristics with few-shot samples. The results are shown in Figs.~\ref{fig:figs-method-results}--\ref{fig:figs-s03}. Compared with the baseline model without FDEP, both \textit{``+ FDEP w/o CO-ISD''} and \textit{``+ FDEP Framework''} produce more complete target structures, clearer boundary contours, and significantly fewer missed and false detections across various scenes. This clearly indicates that the proposed FDEP framework can effectively enhance the model’s sensitivity to small targets and improve its robustness under complex backgrounds and extreme class-imbalance conditions. At the same time, for the same SIRST network, the detection performance using \textit{``+ FDEP w/o CO-ISD''} and \textit{``+ FDEP Framework''} is similar, indicating that the CO-ISD strategy ensures that the lightweight branch can obtain semantic representations similar to the main branch while maintaining lightweight inference, making the detection results stable and reliable. Notably, compared to using \textit{``+ FDEP w/o CO-ISD''}, using \textit{``+ FDEP Framework''} reduces inference time overhead by 40.6\% (from 0.160 to 0.095) to 74.4\% (from 0.082 to 0.021) while maintaining competitive detection accuracy. In summary, the proposed FDEP framework not only significantly enhances the model’s capability in small target discrimination and background suppression under few-shot conditions, but also delivers stable and consistent improvements in both visual quality and quantitative metrics, demonstrating strong potential for practical application.

\section{More Discussion}
\label{sec:more Discussion}
\textbf{\textit{1) Conventional distillation methods fail in SIRST detection tasks.}} According to our experimental observations, when following a traditional model distillation process, that is, when pseudo-labels are generated by the model trained with the \textit{``FDEP w/o CO-ISD''} version and then used to perform fully supervised training on the original SIRST network, the network will collapse directly in the early iteration stage. Even under conditions of joint supervision by real and pseudo labels, the training still collapses. The reason is that SIRST task has characteristics such as extreme class imbalance, small target size and low signal-to-noise ratio. The pseudo-labels generated in the traditional distillation often have semantic noise and confidence bias, which will introduce incorrect gradient signals during backpropagation, leading to instability or even collapse of the network optimization. Furthermore, conventional distillation assumes a stable semantic consistency between the teacher and student networks. However, in SIRST detection, the teacher model’s features are dominated by global semantics derived from the VFMs, while the student network focuses on low-level textures and local contrast modeling. This semantic discrepancy between the two leads to misaligned feature spaces, making direct semantic matching infeasible. Such cross-domain semantic inconsistency causes the explicit distillation loss to act as an optimization disturbance rather than a benefit during training. The proposed CO-ISD strategy fundamentally avoids the instability caused by explicit distillation. Its implicit semantic transfer not only ensures the stability of the training process but also achieves a significant improvement in lightweight branch performance without additional inference costs.

% \begin{table}[t]
% \caption {Computational cost statistics on the SIRST3 dataset. Test time is measured per image.}
% \centering
% \label{tab:tabs-compute-analyse}
% \setlength{\tabcolsep}{1mm}
% \renewcommand{\arraystretch}{1.2}
% \resizebox{1.0\linewidth}{!}{
% \begin{tabular}{c|c|c|c|c|c|c}
% \hline
% Schemes          & Train Time & Train memory & Train FLOPs & Train parameters & Test time   & Test FLOPs \\ \hline
% MSDA-Net         & 11.46 h    & 12.45 G      & 384.88 G    & 4.79 M           & 35 ms & 128.48 G   \\ \hline
% \rowcolor[HTML]{D7D7D7} 
% MSDA-Net w/ FDEP & 19.52 h    & 23.58 G      & 1495.06 G   & 5.24 M           & 35 ms & 128.48 G   \\ \hline
% \end{tabular}
% }
% \vspace{-6pt}
% \end{table}

\begin{table}[t]
\caption {Multi-metric performance comparison on the SIRST3 dataset. Values in parentheses denote absolute \textit{HSE} gains over the corresponding original detector. S@0.1 FPPI and S@1.0 FPPI denote FROC sensitivity at 0.1 and 1.0 false positives per image (FPPI), respectively. nFROC-AUC@1 FPPI denotes normalized FROC-AUC up to 1.0 FPPI.}
\centering
\label{tab:tabs-multi_metric_comparison}
\setlength{\tabcolsep}{1mm}
\renewcommand{\arraystretch}{1.2}
\resizebox{1.0\linewidth}{!}{
\begin{tabular}{c|c|c|c|c|c|c|c}
\hline
Schemes                                                               & HSE $\uparrow$ & hIoU $\uparrow$ & ROC-AUC $\uparrow$ & AP $\uparrow$ & S@0.1 FPPI $\uparrow$ & S@1.0 FPPI $\uparrow$ & nFROC-AUC@1 FPPI $\uparrow$ \\ \hline
ACM (Original)                                                        & 68.05          & 55.05           & 98.69              & 81.86         & 36.11                 & 96.74                 & 87.42                       \\
\rowcolor[HTML]{D7D7D7} 
+ FDEP Framework                                                  & 75.81 \color[HTML]{7030A0}(+7.76)  & 61.99           & 99.65              & 88.98         & 54.90                 & 97.41                 & 90.77                       \\ \hline
ALCNet (Original)                                                     & 74.20          & 62.18           & 99.37              & 86.50         & 69.58                 & 97.01                 & 90.00                       \\
\rowcolor[HTML]{D7D7D7} 
+ FDEP Framework                                               & 78.31 \color[HTML]{7030A0}(+4.11)  & 67.06           & 98.83              & 88.92         & 94.68                 & 96.35                 & 91.56                       \\ \hline
MLCL-Net (Original)                                                   & 86.80          & 75.55           & 98.74              & 93.84         & 96.15                 & 97.87                 & 93.37                       \\
\rowcolor[HTML]{D7D7D7} 
+ FDEP Framework                                             & 87.20 \color[HTML]{7030A0}(+0.40)  & 79.08           & 98.31              & 93.42         & 95.75                 & 97.67                 & 94.46                       \\ \hline
ALCL-Net (Original)                                                   & 86.75          & 77.73           & 99.06              & 94.63         & 96.66                 & 97.81                 & 93.88                       \\
\rowcolor[HTML]{D7D7D7} 
+ FDEP Framework                                             & 87.40 \color[HTML]{7030A0}(+0.65)  & 79.27           & 99.19              & 95.28         & 96.74                 & 97.54                 & 92.16                       \\ \hline
DNA-Net (Original)                                                    & 89.10          & 81.55           & 98.46              & 94.51         & 97.01                 & 98.34                 & 94.97                       \\
\rowcolor[HTML]{D7D7D7} 
+ FDEP Framework                                              & 89.62 \color[HTML]{7030A0}(+0.52)  & 81.38           & 98.49              & 94.53         & 97.80                 & 98.41                 & 95.36                       \\ \hline
GGL-Net (Original)                                                    & 87.99          & 81.11           & 98.83              & 94.62         & 97.94                 & 98.21                 & 95.85                       \\
\rowcolor[HTML]{D7D7D7} 
+ FDEP Framework                                              & 90.69 \color[HTML]{7030A0}(+2.70)  & 82.05           & 99.19              & 95.95         & 97.61                 & 98.07                 & 95.55                       \\ \hline
UIU-Net (Original)                                                    & 88.02          & 79.97           & 99.42              & 95.66         & 96.74                 & 97.08                 & 94.80                       \\
\rowcolor[HTML]{D7D7D7} 
+ FDEP Framework                                               & 90.01 \color[HTML]{7030A0}(+1.99)  & 80.61           & 99.41              & 96.19         & 97.48                 & 97.87                 & 95.50                       \\ \hline
MSDA-Net (Original)                                                   & 89.72          & 80.55           & 99.50              & 96.13         & 97.27                 & 98.07                 & 94.86                       \\
\rowcolor[HTML]{D7D7D7} 
+ FDEP Framework                                             & 91.31 \color[HTML]{7030A0}(+1.59)  & 81.59           & 99.39              & 96.35         & 98.14                 & 98.47                 & 95.93                       \\ \hline
\end{tabular}
}
\vspace{-10pt}
\end{table}

\begin{table}[t]
\caption {Spearman rank correlations of \textit{HSE} with other metrics across 16 configurations in Tab.~\ref{tab:tabs-multi_metric_comparison} on the SIRST3 dataset. S@0.1/1.0 FPPI denotes FROC sensitivity at 0.1/1.0 false positives per image (FPPI). nFROC-AUC@1 FPPI denotes normalized FROC-AUC up to 1.0 FPPI.}
\centering
\label{tab:tabs-metric_rank_correlation}
\setlength{\tabcolsep}{3mm}
\renewcommand{\arraystretch}{1.2}
\resizebox{1.0\linewidth}{!}{
\begin{tabular}{c|c|c|c|c|c|c}
\hline
Compared Metric & hIoU  & ROC-AUC & AP    & S@0.1 FPPI & S@1.0 FPPI & nFROC-AUC@1 FPPI \\ \hline
Spearman $\rho$ & 0.935 & 0.191   & 0.879 & 0.904      & 0.786      & 0.909            \\ \hline
\end{tabular}
}
\vspace{-15pt}
\end{table}

\textbf{\textit{2) Rationale and advantages of the HSE.}} Existing single metrics are insufficient to comprehensively characterize detection capability, and the current evaluation system is highly sensitive to threshold settings. To overcome the fragmentation and insufficiency of existing evaluation systems, we construct a Holistic SIRST Evaluation (\textit{HSE}) metric, which performs multi-threshold integral assessment at both pixel-level confidence prediction and target-level robust detection perspectives, providing a more stable and more comprehensive criterion for fair model comparison and performance analysis. Notably, the proposed \textit{HSE} fundamentally differs from \textit{hIoU}~\cite{pang2025rethinking}. The \textit{hIoU} adopts hierarchical discrete decisions and emphasizes target-level results by considering pixel-level IoU only for correctly detected targets. In contrast, our \textit{HSE} unifies pixel- and target-level evaluation via continuous threshold integration, jointly capturing pixel-level confidence stability and target-level robustness while avoiding single-threshold bias. As shown in Fig.~\ref{fig:figs-roc} of the Main Text and Tab.~\ref{tab:tabs-s05} of the Appendix, \textit{hIoU} has limited discriminative power both between and within sample pairs. For example, Case 2-S1 and Case 2-S2 differ in false positive area, but their \textit{hIoU} scores are the same. Our \textit{HSE} can more reasonably distinguish these differences. In addition, \textit{HSE-P}, a component of \textit{HSE}, can alleviate the limitations of the ROC curve. From the  Fig.~\ref{fig:figs-roc} of the Main Text and the Tab.~\ref{tab:tabs-s05} of the Appendix, the AUC of ROC curves in multiple cases does not faithfully reflect true pixel-level performance. \textit{HSE-P} mitigates this issue. For instance, compared with Case 1-S2, Case 1-S1 contains severe false positives but still achieves a higher AUC. For \textit{HSE-P}, Case 1-S1 has a lower value,  which better aligns with the actual pixel-level quality.

% \textbf{\textit{3) Computational cost analysis.}} We provide detailed computational cost statistics on the SIRST3 dataset. From Tab.~\ref{tab:tabs-compute-analyse}, FDEP introduces some manageable additional overhead during the training phase, primarily from feature extraction of the frozen VFM, SAMF fusion, and CO-ISD dual-branch optimization. FDEP is not designed to reduce training costs, but rather to transfer the global semantic priors provided by the VFM to a lightweight SIRST network with acceptable training overhead, thereby improving detection performance without introducing additional inference costs. It's worth emphasizing that the VFM, SAMF, and auxiliary branches are removed during the inference phase. Therefore, the inference time and FLOPs remain consistent with the original network. Given that the training cost is a one-time expense, while inference efficiency directly impacts actual deployment, the increased training overhead is reasonable and acceptable.

\textbf{\textit{3) Comparison with existing evaluation metrics.}} To systematically examine the relationship between \textit{HSE} and existing evaluation metrics, we compare 16 configurations on the SIRST3 dataset, covering holistic metrics (\textit{HSE} and hIoU), pixel-level metrics (AP and ROC-AUC), and FROC-style target-level metrics. These configurations comprise eight backbone detectors and their corresponding FDEP-equipped variants. First, \textit{HSE} shows high overall ranking consistency with existing metrics. From Tab.~\ref{tab:tabs-metric_rank_correlation}, except for ROC-AUC, the Spearman correlations between \textit{HSE} and the other metrics range from 0.786 to 0.935, indicating broad agreement with existing holistic, pixel-level, and target-level metrics across the evaluated configurations. In contrast, the correlation between \textit{HSE} and ROC-AUC is only 0.191. For example, ACM equipped with FDEP achieves the highest ROC-AUC (99.65) but substantially lower AP, FROC, and \textit{HSE} scores than several other configurations, further supporting our discussion regarding the limitations of ROC-AUC in SIRST evaluation. Second, local ranking differences are interpretable. \textit{HSE} jointly considers pixel-level confidence quality and target-level detection robustness, so its ranking need not be identical to that of any individual metric. From Tab.~\ref{tab:tabs-multi_metric_comparison}, ``GGL-Net + FDEP Framework'' achieves a slightly higher hIoU than ``MSDA-Net + FDEP Framework'' (82.05 vs.\ 81.59), whereas the latter obtains higher AP and FROC scores, while also achieving a higher \textit{HSE} (91.31 vs.\ 90.69). This difference reflects their distinct evaluation criteria: hIoU relies on hierarchical discrete decisions, whereas \textit{HSE} unifies pixel- and target-level evaluation through multi-threshold integration, jointly capturing pixel-level confidence quality and target-level robustness while reducing single-threshold bias. In addition, the overall effectiveness of FDEP is not solely supported by \textit{HSE}. FDEP improves \textit{HSE} on all eight backbones, hIoU and AP on seven, Sensitivity@0.1 FPPI and normalized FROC-AUC@1 FPPI on six, and Sensitivity@1.0 FPPI on four. These results indicate that the overall advantage of FDEP is also supported by multiple existing evaluation metrics, although improvements are not uniform across all metrics and backbones. Overall, \textit{HSE} shows strong ranking agreement with hIoU, AP, and FROC-style metrics, while providing reasonable differences when pixel- and target-level performance are imbalanced.

\section{Broader Impacts}
\label{sec:broader_impacts}

This paper focuses on SIRST detection, aiming to improve detection accuracy while maintaining inference efficiency by introducing frozen VFM representations. The proposed FDEP framework utilizes global semantic priors during the training phase and retains only lightweight branches during the inference phase, contributing to efficient and robust infrared perception. Related technologies can serve practical applications such as maritime monitoring, autonomous driving, environmental perception, and early warning, enhancing the detection capabilities of weak and distant targets in complex backgrounds. We hope this work will promote the application of VFMs in SIRST detection and provide technical support for the efficient deployment and reliable application of related methods.

\section{Limitations and Future Work}
\label{sec:limitations_and_future_work}

Although the proposed FDEP framework shows strong compatibility, can be seamlessly adapted to existing SIRST detection networks, and achieves stable performance improvements on multiple public datasets, it still has several limitations. On the one hand, although this work has been extensively validated on multiple public datasets, the available data are still limited and do not fully cover complex real-world deployment scenarios, such as extreme weather and low-quality imaging. Therefore, its generalization ability in broader practical environments requires further evaluation. On the other hand, FDEP relies on VFMs to provide global semantic priors during training, which introduces additional training cost. According to our statistics, on the largest SIRST3 dataset, the training times of the original MSDA-Net and MSDA-Net equipped with FDEP are 11.46 h and 19.52 h, respectively. Nevertheless, it is worth emphasizing that FDEP retains only the lightweight branch during inference, and thus does not introduce additional inference-time overhead. As a result, the two models maintain the same inference speed in practical applications. Future work will further explore more efficient VFMs adaptation and feature compression strategies, and construct more comprehensive real-world datasets to improve training efficiency, generalization ability, and deployment reliability.

%%%%%%%%%%%%%%%%%%%%%%%%%%%%%%%%%%%%%%%%%%%%%%%%%%%%%%%%%%%%

% \newpage
% \input{checklist.tex}

\end{document}